\documentclass[conference]{IEEEtran}

\usepackage{cite}
\usepackage{amsmath,amssymb,amsfonts}
\usepackage{amsthm}
\newtheorem{definition}{Definition}
\usepackage{algorithmic}
\usepackage{graphicx}
\usepackage{textcomp}
\usepackage{xcolor}
\usepackage[ruled]{algorithm2e}
\usepackage{caption}
\usepackage{subcaption}
\usepackage{multirow}
\usepackage{booktabs}
\newcommand{\ra}[1]{\renewcommand{\arraystretch}{#1}}

\def\BibTeX{{\rm B\kern-.05em{\sc i\kern-.025em b}\kern-.08em
    T\kern-.1667em\lower.7ex\hbox{E}\kern-.125emX}}
\begin{document}

\title{Empirically Measuring Transfer Distance for System Design and Operation}

\author{\IEEEauthorblockN{Tyler Cody}
\IEEEauthorblockA{\textit{Engineering Systems and Environment} \\
\textit{University of Virginia}\\
Charlottesville, USA \\
}
\and
\IEEEauthorblockN{Stephen Adams}
\IEEEauthorblockA{\textit{Engineering Systems and Environment} \\
\textit{University of Virginia}\\
Charlottesville, USA \\
}
\and
\IEEEauthorblockN{Peter A. Beling}
\IEEEauthorblockA{\textit{Engineering Systems and Environment} \\
\textit{University of Virginia}\\
Charlottesville, USA \\
}
}

\maketitle

\begin{abstract}
Classical machine learning approaches are sensitive to non-stationarity. Transfer learning can address non-stationarity by sharing knowledge from one system to another, however, in areas like machine prognostics and defense, data is fundamentally limited. Therefore, transfer learning algorithms have little, if any, examples from which to learn. Herein, we suggest that these constraints on algorithmic learning can be addressed by systems engineering. We formally define \emph{transfer distance} in general terms and demonstrate its use in empirically quantifying the transferability of models. We consider the use of transfer distance in the design of machine rebuild procedures to allow for transferable prognostic models. We also consider the use of transfer distance in predicting operational performance in computer vision. Practitioners can use the presented methodology to design and operate systems with consideration for the learning theoretic challenges faced by component learning systems.
\end{abstract}

\begin{IEEEkeywords}
transfer learning, prognostics, computer vision
\end{IEEEkeywords}

\section{Introduction}

Machine learning is moving from laboratories to the field, however, the identically distributed environments found in the lab are rarely found in the real-world. Algorithmic approaches for dealing with non-stationarity rely heavily on data from the new environment, however, such data is not always available.

Applied machine learning for prognostics and health management (PHM) is prototypical of this trend and challenge. Non-stationarities are unavoidable in PHM for machinery. Differences in manufacturing and installment give supposedly identical machines different initial conditions, and phenomena such as degradation, repair, and part replacement cause behavior to drift over a machine's life cycle. Adding to these challenges, labeled data from fielded machines is rarely available because when a failure occurs, the machine is repaired or rebuilt, inducing a distribution change, or rendered irreparable. 

Similarly, in defense settings, imagery related to new missions is limited. Data collection for new missions is costly. It can require operating in hostile territory or airspace and within enemy field of fire. Moreover, battlefields are dynamic and often do not afford data collection at the scale required by existing data-driven computer vision methods. Additionally, defense is game-theoretic in nature, and adversaries can manipulate the appearance of concerns such as aircraft or ground vehicles to take advantage of an over-reliance on data \cite{rogers2019adversarial}.

In both PHM and defense, algorithmic approaches for relating behaviors between systems and over time are fundamentally constrained. Instead of focusing on engineering ever-more adaptive learning systems, we suggest a focus on methodologies that support the design and operation of systems to limit non-stationarities to acceptable levels. This interdisciplinary approach treats generalization, i.e., satisfactory predictive performance on new data, as a systems-level goal, not a goal exclusive to algorithm design.

Designing and operating in this way requires metrics that bring the learning theoretic challenges of learning systems to the systems-level. \emph{Transfer distance}, the abstract distance knowledge must traverse to transfer from one system to another, is focal in domain adaptation theory and is used to relate the magnitude of distributional change between systems to prediction error in the new system. Although transfer distance is typically left as an informal notion or implicit in transfer learning methods, here, we formalize it and position it as central to the characterization of the relationship between systems and the generalization of their component learning systems.

We present a Bayesian approach for empirically quantifying transfer distance. The accompanying studies offer a guide for practitioners on how to quantify the difficulty of transfer, the transferability of different learning tasks, and the role of sample size in transferability, as well as how to use transfer distance to quantify expected operational performance. We consider a case in hydraulic actuator health monitoring where non-stationarities occur as the result of actuator rebuilds. We also consider a case in computer vision with a mission context, where information regarding a mission's expected operating environment is used to assess expected operational performance. We frame the former in terms of system design and the latter in terms of system operation. In doing so, we contribute to the broader effort of developing principled methodologies for the systems engineering of artificial intelligence (AI).

This paper is structured as follows. First, we provide background on transfer learning, domain adaptation, concept drift, PHM, and computer vision. We then justify the use of transfer distance as a metric by drawing from domain adaptation theory and present our methodology for quantifying transfer distance. Subsequently, we apply our methodology to characterize the transfer learning problems induced by an actuator rebuild procedure and mission deployment. We conclude with a synopsis and a statement of future work.

\section{Background}

We briefly review transfer learning, domain adaptation, concept drift, PHM, and computer vision, and note this paper's relationship to them. In short, this paper presents PHM and computer vision case studies in empirically characterizing transferability using principles from domain adaptation and methods from concept drift.

\subsection{Transfer Learning}

Transfer learning describes the idea of using knowledge from source systems to help learn in a particular target system. More formally, consider a learning problem that consists of a domain $\mathcal{D}=\{ \mathcal{X}, P(X) \}$ and a learning task $\mathcal{T}=\{ \mathcal{Y}, P(Y|X) \}$, where $X = x$ for $x \in \mathcal{X}$ and $Y = y$ for $y \in \mathcal{Y}$. Transfer learning uses knowledge from a source learning problem $\{ \mathcal{D}_S, \mathcal{T}_S \}$ to improve the learning of a function $f_T:\mathcal{X}_T \to \mathcal{Y}_T$ in a target learning problem $\{ \mathcal{D}_T, \mathcal{T}_T \}$ where $\mathcal{D}_S \neq \mathcal{D}_T$ or $\mathcal{T}_S \neq \mathcal{T}_T$ \cite{pan2009survey}.

Transfer learning enables learning in environments where data is limited. Perhaps more importantly, it allows learning systems to propagate their knowledge forward through distributional changes, such as the degradation and wear of physical components, changes in use cases and functionality, and policy changes regarding the use of particular features $\mathcal{X}$ and labels $\mathcal{Y}$ \cite{cody2019systems}. The classical approaches to transfer learning involve selecting or weighing samples from the source, projecting the source and target features into a latent space, or bounding the parameters of the target model within a range of the source model's parameters \cite{pan2009survey}.

Identifying whether or not transfer learning is an appropriate solution for a particular learning problem is crucial \cite{rosenstein2005transfer}. Failure to do so can result in \emph{negative transfer}, wherein dissimilarity between the source and target systems results in transfer learning under-performing traditional machine learning approaches. While the extent of negative transfer is algorithm-dependent, the existence of negative transfer is tied to the distributions underlying the learning problem \cite{wang2019characterizing}. Thus, closeness between the source and target distributions is a pre-condition for transfer learning success.

\subsection{Domain Adaptation}

Domain adaptation is a sub-field of transfer learning where $\mathcal{X_S} \times \mathcal{Y_S} = \mathcal{X_T} \times \mathcal{Y_T}$ \cite{jiang2008literature}. In other words, only the probability distributions change between the sources and target, not their sample spaces. Domain adaptation theory places transfer distance at the center of bounding error in new environments \cite{blitzer2008learning, ben2010theory}. The common approach taken is to note that the error in the target is related to the error in the source plus some measure of similarity between the source and target.

These bounds can be loosely represented by the following inequality,
\begin{equation}
    \epsilon_T \leq \epsilon_S + \delta + C
    \label{eq:1}
\end{equation}

\noindent where $\epsilon_T$ and $\epsilon_S$ are the errors in the source and target, $\delta$ is the transfer distance, and $C$ is a constant term which accounts for relevant complexities, for example, VC-dimension \cite{ben2010theory}. Although Inequality \ref{eq:1} is an approximation of the underlying learning theory, specifications can be added to arrive at proper, learning theoretic bounds using statistical divergence \cite{blitzer2008learning}, $\mathcal{H}$-divergence \cite{ben2010theory}, Rademacher complexity \cite{mohri2009rademacher}, or integral probability metrics \cite{zhang2012generalization}.

\subsection{Concept Drift}

Whereas transfer learning considers distributional change between a source and target, concept drift considers distributional change that occurs in streaming data from one stable distribution, termed a concept, to another. There are many metrics similar to transfer distance used in the concept drift literature to characterize drift \cite{webb2016characterizing}. Drift in these streaming systems has been modeled and simulated using Gaussian mixture models \cite{wu2005tracking, diaz2018clustering}. Many methods use Hellinger distance to calculate distributional divergence because it is bounded $[0, 1]$ and symmetric \cite{ditzler2011hellinger, webb2016characterizing}. Consistent with concept drift literature, we use a combination of Gaussian mixture models and Hellinger distance to characterize distributional change.

\subsection{PHM}

\begin{figure}[t]
    \centering
    \includegraphics[width=0.4\textwidth]{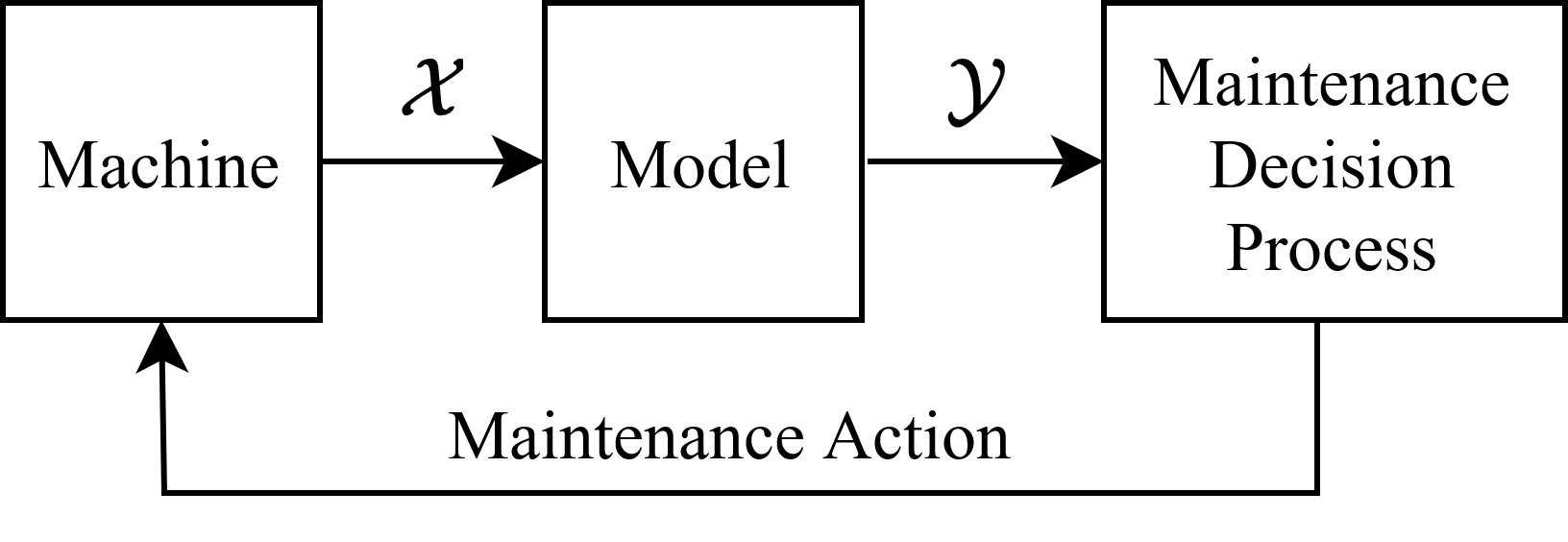}
    \caption{Data-driven models inform maintenance actions which change the distribution of their data. Non-stationarity is inherent in PHM.}
    \label{fig:maintenance}
\end{figure}

PHM is concerned with the use of prognostics and diagnostics for the management of machine health \cite{tsui2015prognostics}. In mechanized systems generally, it is essential for continuous operation, and, thus, is an important field of engineering research. As machine down-time is the eminent failure in production systems \cite{li2008production}, PHM is crucial to economic productivity. Furthermore, PHM helps safeguard critical systems such as gears in rotorcraft \cite{delgado2012survey}, whose failure can cause loss of propulsion mid-flight, and air filtration systems \cite{landolsi2017air}, whose failure in high pressure environments such as submarines can be equally catastrophic, among others \cite{eftekhari2013online}.

Currently, machine health management is dominated by time-based maintenance schedules, however, there is an increasing interest in and use of data-driven PHM for adaptive scheduling \cite{zhang2019data}. This has led to extensive application of machine learning for health state classification and remaining useful life regression. There is a much smaller body of literature, however, using transfer learning to deal with the challenges these methods face in practice due to the aforementioned non-stationarities and label constraints \cite{shen2015bearing, xie2016cross, lu2017deep, zhang2017new, li2020fault}.

Non-stationarity is a fundamental challenge in PHM. In data-driven PHM, sensor data from machines is used for prognostics and diagnostics to inform operations management. When a maintenance action is taken, such as a machine rebuild, where the machine is deconstructed and rebuilt, the distribution of the sensor data changes. This cycle is represented in Figure \ref{fig:maintenance}. Minor physical differences in the tensions of fasteners or locations of sensors can degrade predictive performance. The extent of degradation is difficult to ascertain because after an example of failure occurs, the system will be repaired, inducing a distribution change, or will be deemed irreparable.

Thus, in PHM systems, there is a real limit in our ability to address non-stationarity with algorithm design; it is necessary to take into account the role of system design in the generalization of learning. And to that end, it is necessary to have metrics which can link notions like the design of maintenance procedures, e.g., regarding details like tensions and sensor locations, to notions like the transferability of knowledge.

In recent work, we extensively studied PHM for hydraulic actuators in order to better place related data-driven modeling in a systems context, including cost and power constraints \cite{adams2017comparison, meekins2018cost, farinholt2018developing,adams2019hierarchical}. These studies have used a fault-simulating test-bed that consists of two matched rotary actuators, where one acts as the actuator and the other acts as the load \cite{adams2016condition}. Here, we use data collected from this test-bed to extend the literature on data-driven PHM for hydraulic actuators by explicitly modeling the transfer distance associated with a rebuild procedure. Previously, we showed that sample transfer can be used to recover performance across the rebuild \cite{cody2019transferring}. Here, instead of solving the transfer learning problem, in contrast, we use transfer distance as a means of characterizing the transfer learning problem associated with the rebuild.

\subsection{Computer Vision}

Computer vision is a broad field concerned with visual perception and pattern recognition. In recent years, deep learning has overtaken handcrafted feature engineering methods for processing images in the computer vision research literature \cite{xiao2016overview, nanni2017handcrafted}. Instead of extracting expert-defined features from images as a pre-processing step, deep learning takes raw images as inputs and learns to both extract its own features and make predictions as part of a single, end-to-end process. While deep learning increases predictive performance and allows for novel use cases, it is heavily reliant on large data sets \cite{goodfellow2016deep}. 

As previously described, in defense applications, this presents a bottleneck to deployment. Image classifiers have been trained to detect planes and their orientation when parked in airports' aprons using knowledge transferred from general visual recognition tasks \cite{chen2018end}. However, such models are highly dependent on the airports included in training. As we will demonstrate, classifiers can suffer a decrease in performance when the biomes surrounding the airports change between training and operation. We calculate the transfer distance associated with transferring a model from one geographical region to another, as in a mission deployment scenario, and use it to anticipate model degradation. We use an auto-encoder for dimension reduction, similar to existing approaches to explainable AI \cite{bulusu2020anomalous}.

\section{Methods}

Transfer distance is usually referred to informally, e.g., to describe \emph{near} or \emph{far} transfer. It is implicit in the use of Wasserstein distance \cite{shen2017wasserstein}, maximum mean discrepancy \cite{pan2008transfer, long2017deep}, generative adversarial networks \cite{tzeng2015simultaneous, ganin2016domain}, and others, to calculate distributional-divergence-based components of loss functions in transfer learning algorithms. We consider transfer distance explicitly, in a way that may not necessarily be useful in calculating loss functions, but is interpretable to system designers and operators. Transfer distance is defined as follows.

\begin{definition}{\emph{Transfer distance.}}\\
The transfer distance between a source and target learning system, denoted $S$ and $T$ respectively, is a measure $\delta$,
$$\delta: P_S \times P_T \to {\rm I\!R}$$
on probability measures $P_S$ and $P_T$ from the source domain $\mathcal{D}_S$ and task $\mathcal{T}_S$ and target domain $\mathcal{D}_T$ and task $\mathcal{T}_T$, respectively.
\end{definition}

More general definitions are possible. This definition directs our interest towards the marginal distributions $P(X)$ from the domains $\mathcal{D}$ and posterior distributions $P(Y|X)$ from the tasks $\mathcal{T}$. For the purposes of explainability and analysis, we model these distributions explicitly, in closed-form, and take a Bayesian approach to constructing the posterior. We only fit $P(X|Y)$, and, using an estimate for the prior $P(Y)$, construct the marginal $P(X)$ and posterior $P(Y|X)$.

Our algorithm for computing transfer distances can be described as follows. We assume that $\mathcal{X}_S=\mathcal{X}_T=\mathcal{X}$, $\mathcal{Y}_S=\mathcal{Y}_T=\mathcal{Y}$, that $\mathcal{X}$ is continuous, and that $\mathcal{Y}$ is discrete. We first fit the likelihood distributions $P_S(X|Y=y)$ and $P_T(X|Y=y)$ for all $y \in \mathcal{Y}$. We construct $P_S(X)$ and $P_T(X)$ using a prior $P(Y)$ and the total probability law, and then construct $P_S(Y=y|X)$ and $P_T(Y=y|X)$ for all $y \in \mathcal{Y}$ using Bayes theorem. We then sample from $\mathcal{X} \times \mathcal{Y}$ according to the source and target distributions and calculate the transfer distance $\delta$ using these samples. This process is shown in Algorithm \ref{algo:td}.

\begin{algorithm*}[t]
\DontPrintSemicolon
\KwIn{$data_S$, $data_T$, $P(Y=y) \forall y \in \mathcal{Y}$}
\KwOut{$\delta_{X|Y=y}, \delta_{X}, \delta_{Y=y|X}$}
\SetKwProg{Def}{def}{:}{}
\Def{\emph{fit}(data)}{
    $\{P(X|Y=y)\}_{y\in\mathcal{Y}} \gets$ fitter$(data)$\;
    $P(X) \gets \sum_{y \in \mathcal{Y}} P(X|Y=y)P(Y=y)$\;
    $\{P(Y=y|X)\}_{y\in\mathcal{Y}} \gets \{P(X|Y=y)P(Y=y)/P(X)\}_{y \in \mathcal{Y}}$\;
    return $P(X|Y), P(X), P(Y|X)$\;
}
\;
$\{P_S(X|Y=y)\}_{y\in\mathcal{Y}}, P_S(X), \{P_S(Y=y|X)\}_{y\in\mathcal{Y}} \gets$ fit$(data_S)$\;
$\{P_T(X|Y=y)\}_{y\in\mathcal{Y}}, P_T(X), \{P_T(Y=y|X)\}_{y\in\mathcal{Y}} \gets$ fit$(data_T)$\;
\;
$\{\delta_{X|Y=y}\}_{y\in\mathcal{Y}} \gets \delta(P_S(X|Y), P_T(X|Y))$\;
$\delta_{X} \gets \delta(P_S(X), P_T(X))$\;
$\{\delta_{Y=y|X}\}_{y\in\mathcal{Y}} \gets \delta(P_S(Y|X), P_T(Y|X))$\;
\;
return $\{\delta_{X|Y=y}\}_{y\in\mathcal{Y}}, \delta_{X}, \{\delta_{Y=y|X}\}_{y\in\mathcal{Y}}$\;
\caption{Calculating Transfer Distance in Domain Adaptation with Discrete $\mathcal{Y}$}
\label{algo:td}
\end{algorithm*}

We use Gaussian mixture models (GMMs) to fit the likelihoods $P(X|Y)$, i.e., as the \emph{fitter} method in \emph{fit} function of Algorithm \ref{algo:td}. Gaussian mixture modeling is a clustering technique whereby a mixture of probability weighted multi-variate Gaussian distributions is fit to data. Each point is assigned to a single multi-variate Gaussian, i.e., its cluster. For a GMM with $K$ clusters,
$$p(X) = \sum_{k=1}^{K}\pi_k \mathcal{N}(X|\mu_k, \sigma_k),$$
where $p(X)$ is the density function of $X$, $\pi_k$ is the probability weight of cluster $k$, and $\mathcal{N}$ is the multi-variate Gaussian distribution with mean $\mu_k$ and co-variance $\sigma_k$.

Explicit, closed-form models of the source and target allow for a rich set of distance functions. Different applications may call for different distances, and closed-form distributions afford this flexibility. In our case, we use the Hellinger distance and Kullback-Leibler (KL) divergence as our transfer distances $\delta$. Given two discrete probability distributions $P=(p_1, ..., p_n)$ and $Q=(q_1, ..., q_n)$, the Hellinger distance between $P$ and $Q$ is
$$H(P, Q) = \frac{1}{\sqrt{2}}\sqrt{\sum_{i=1}^n (\sqrt{p_i} - \sqrt{q_i})^2 }.$$
$H$ is symmetric and bounded $[0, 1]$, where $H=0$ implies that the distributions are completely identical and $H=1$ implies that they do not overlap at all. The KL divergence between $P$ and $Q$ is
$$KL(P, Q) = \sum_{i=1}^n p_i \log{\frac{p_i}{q_i}}.$$
$KL$ is not symmetric and is unbounded above $[0, \infty)$, where its lower bound implies that the distributions are completely identical.

\section{Transfer Distance for System Design}

In system design, transfer distance can be used to design systems with an awareness of the generalization difficulty faced by component learning systems. Generalization difficulty concerns the difficulty of achieving a certain level of error on new data. Different design decisions can be associated with different generalization difficulties. Inequality \ref{eq:1},
$$\epsilon_T \leq \epsilon_S + \delta + C,$$
suggests that a higher transfer distance $\delta$ is associated with a higher demand on the error in the source $\epsilon_S$ and the constant term $C$ to keep the upper bound on error in the target $\epsilon_T$ the same as with a lower transfer distance. Therefore, transfer distance has a strong, fundamental influence on generalization difficulty.

In cases where the distance between the source and target is, for example, associated with some physical change in the system, we can use transfer distance as a means of associating the physical change with generalization difficulty. Consider the generalization of prognostics models across system rebuilds. In previous work, we found that while binary health states for hydraulic actuators can be classified with an accuracy of 98\% when trained and tested on the same actuator, when the actuator is deconstructed and rebuilt, the same classifier does marginally better than random guessing \cite{cody2019transferring}. When transfer learning is applied classification accuracy recovers to almost 90\%. 

In the following, transfer distance is used to characterize the generalization difficulty associated with a particular actuator rebuild procedure. We show how an analysis of transfer distance can be used to understand why the original classifier failed, to suggest why transfer learning worked, and, ultimately, to inform the iterative design of rebuild procedures to limit degradation in predictive performance across system rebuilds. We quantify the generalization difficulty associated with the rebuild procedure in terms of the transfer distance between binary and multi-class health state classification before and after the rebuild. Then, we quantify the number of samples required to achieve a stable estimate of transfer distance.

Faults were simulated on a hydraulic actuator, the actuator was deconstructed and rebuilt, and the faults were re-simulated. The failure modes considered are opposing load, external load, bypass valve, and leak valve failures, among miscellaneous others. The hydraulic actuator test stand is equipped with sensors to collect acceleration, pressure, flow, temperature, and rotary position. In pre-processing, to capture aspects of time-dependence, the data is first windowed and summarized by the mean and standard deviation of each window. Then, to reduce the dimension of the data, principal component analysis is applied. The first two principal components capture 90\% of the variance in the windowed features. These two components are used in our studies.

\subsection{Transfer Distance Induced by Rebuild}

\begin{figure*}[t]
     \centering
     \begin{subfigure}[b]{0.32\textwidth}
         \centering
         \includegraphics[width=\textwidth]{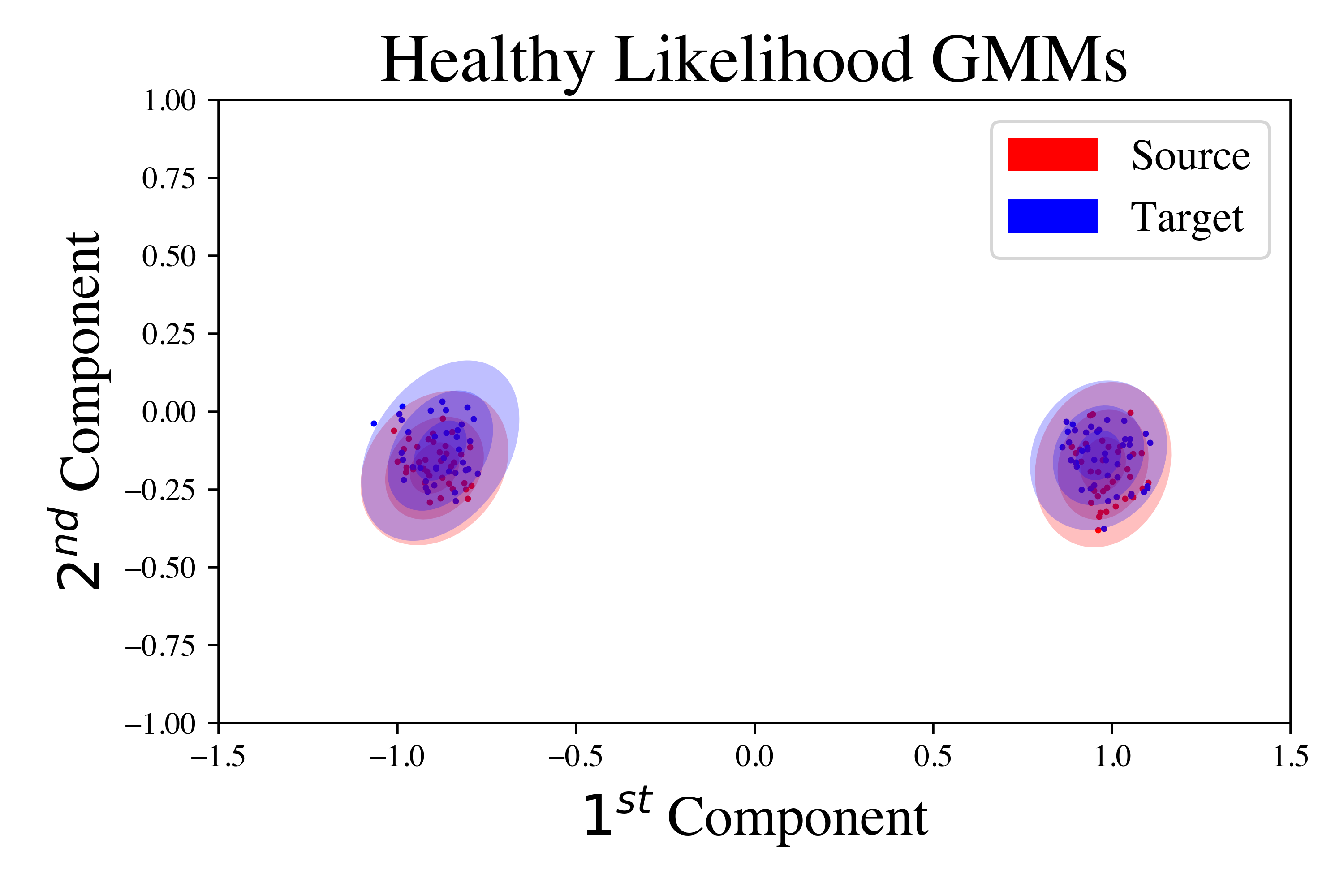}
         \caption{$p(X|Y=0)$}
         \label{fig:healthy-likelihood}
     \end{subfigure}
     \hfill
     \begin{subfigure}[b]{0.32\textwidth}
         \centering
         \includegraphics[width=\textwidth]{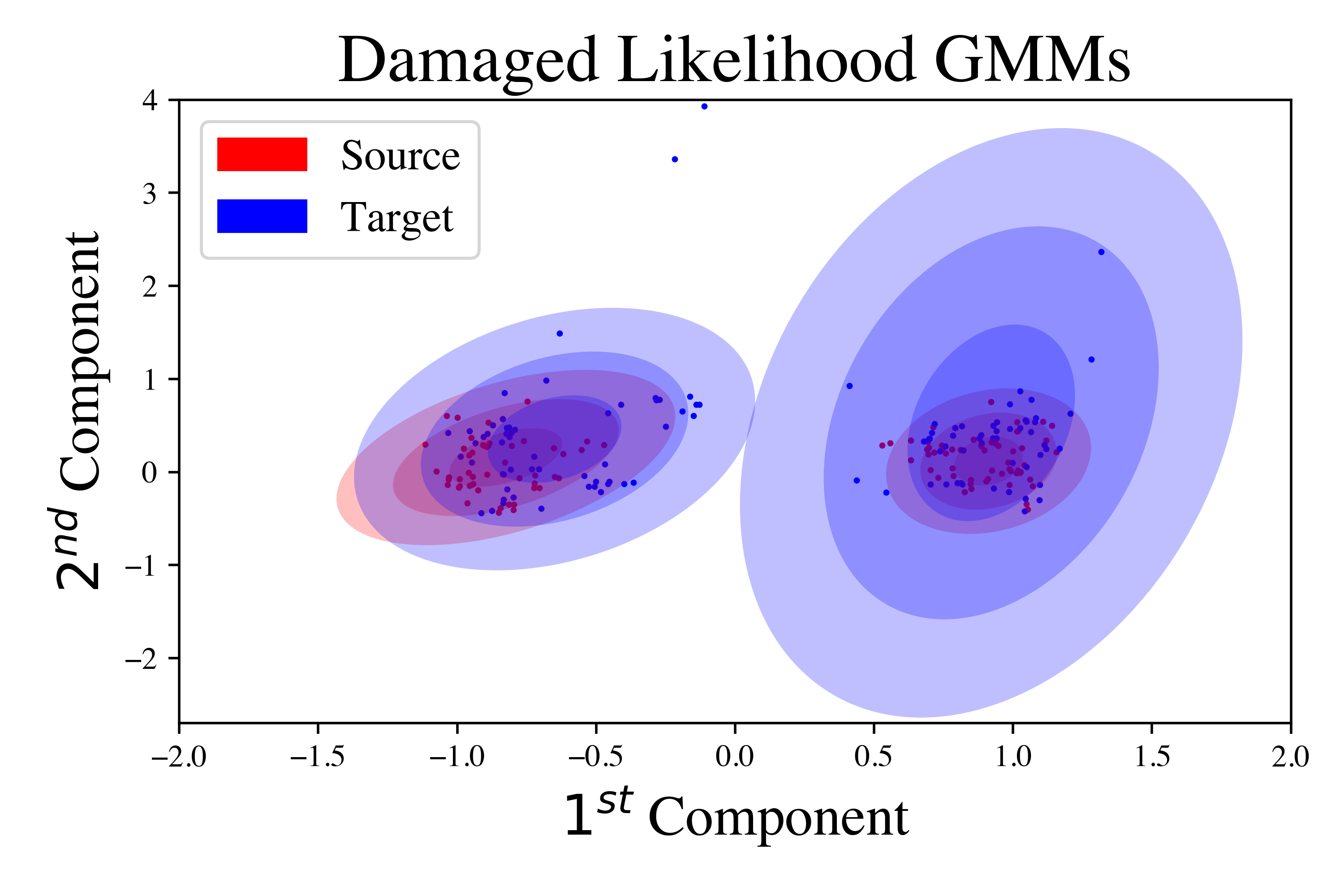}
         \caption{$p(X|Y=1)$}
         \label{fig:damaged-likelihood}
     \end{subfigure}
     \hfill
     \begin{subfigure}[b]{0.32\textwidth}
         \centering
         \includegraphics[width=\textwidth]{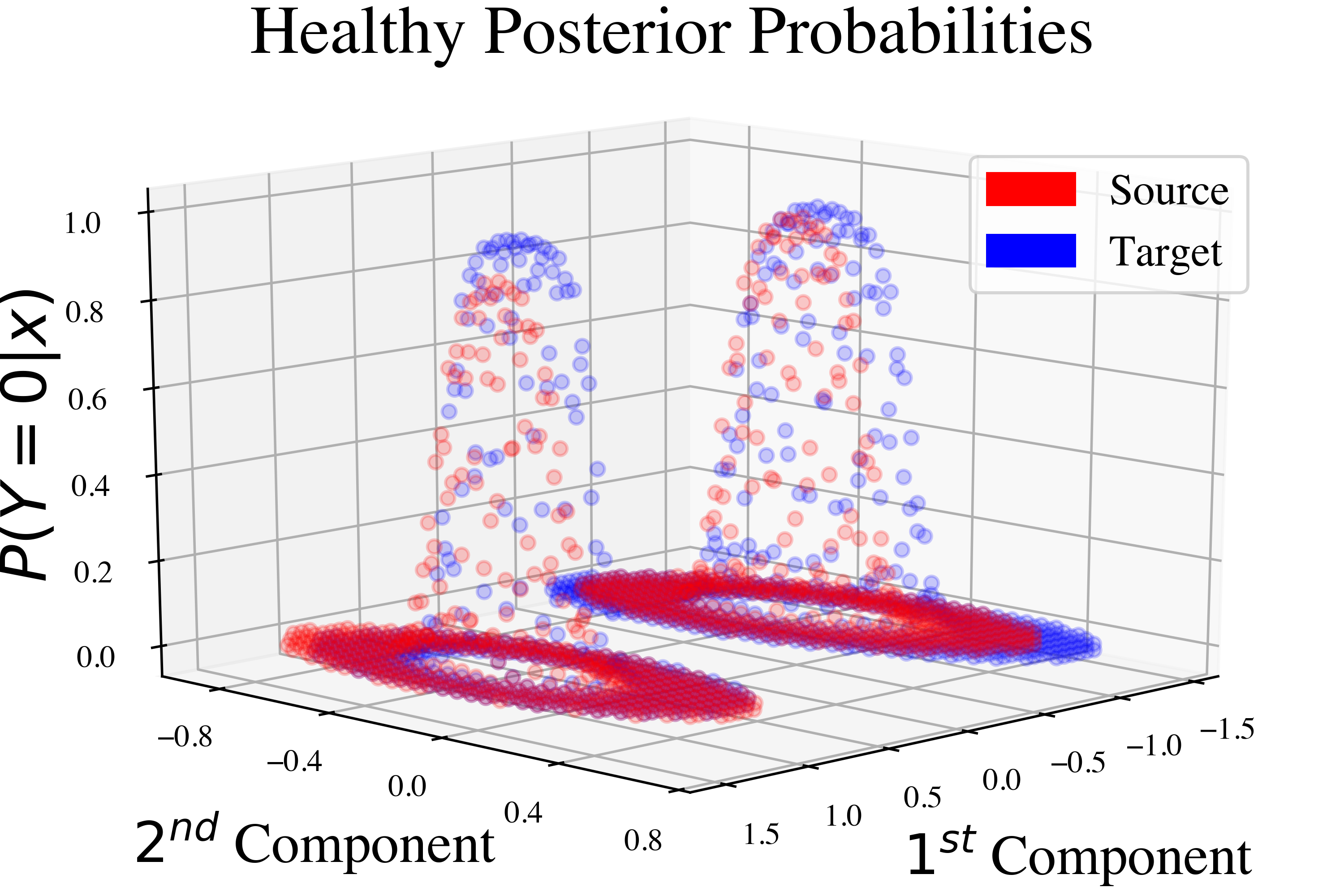}
         \caption{$P(Y=0|X)$}
         \label{fig:posterior}
     \end{subfigure}
        \caption{Likelihood Densities and Healthy Posterior Distributions}
        \label{fig:three-graphs}
\end{figure*}

First, we consider binary health state classification, where we learn to predict whether the hydraulic actuator is healthy, $Y=0$, or damaged, $Y=1$. The original actuator is the source, the rebuilt actuator is the target, and we are interested in empirically quantifying the change in the binary classification problem induced by the rebuild process, i.e., the changes in the distributions underlying the problem. There are 789 healthy and 1480 damage samples in the source, and 1098 and 1822 in the target, respectively.

The empirical prior $P(Y=0)$ is given by the ratio of healthy samples to damaged samples, but such a prior implies almost even odds of failure. We approximate the empirical prior as $P(Y=0)=0.40$ and compare against $P(Y=0) \in \{0.9, 0.99, 0.999\}$. The same priors are used for both the source and target.

The fitted likelihoods and the constructed posterior probability of being healthy are plotted in Figure \ref{fig:three-graphs}. The likelihood densities in Figures \ref{fig:healthy-likelihood} and \ref{fig:damaged-likelihood} show the source in red and target in blue, fit with 2-component Gaussian mixture models, where each concentric ellipse represents 1 standard deviation from a component's mean. The plotted points are from samples held-out from the fitting process. Whereas the healthy densities overlap closely between the source and target, the damaged densities do not. The target, rebuilt actuator has a larger spread in the distribution of damaged data when represented by its first two principal components. Classification likely dropped because of this increased variance. Despite this difference, the posteriors, shown in Figure \ref{fig:posterior} for $P(Y=0)=0.40$, are fairly similar. Transfer learning likely succeeded at bringing accuracy back to nearly 90\% because the increased variance in the damaged likelihood did not strongly affect the posterior.

Transfer distances $\delta$ are shown in Table \ref{table:binary}. As in the plots, the healthy likelihoods are closer than the damaged likelihoods. Notably, the transfer distance between the marginals $P(X)$ is larger than that between the posteriors $P(Y=0|X)$. In other words, there are changes in the distribution of the sensor data that do not have a material effect on the binary classification problem. We can also note that as the the prior odds of failure decrease, $\delta_X$ and $\delta_{Y=0|X}$ decreases as well, because the difference in the damaged likelihood is weighted less.

\begin{table}[h]
\centering
\ra{1.3}
\begin{tabular}{@{}lrrrr@{}}
\toprule
\multirow[t]{2}{*}{Transfer Distance} &
\multicolumn{4}{c}{$P(Y=0)$} \\
\cmidrule{2-5}
& 0.40 & 0.90 & 0.99 & 0.999 \\
\midrule
$\delta_{X|Y=0}$ & 0.22 & - & - & - \\
$\delta_{X|Y=1}$ & 0.54 & - & - & - \\
$\delta_{X}$ & 0.41 & 0.25 & 0.22 & 0.22 \\
$\delta_{Y=0|X}$ & 0.24 & 0.23 & 0.23 & 0.23 \\
\bottomrule
\\
\end{tabular}
\caption{Hellinger transfer distance for relevant distributions. Note, $\delta_{X|Y}$ does not depend on $P(Y)$.}
\label{table:binary}
\end{table}

These results show that the rebuild procedure affects the distributions of damaged data far more than the distribution of healthy data. This means that while healthy behavior appears similar across rebuilds, failure does not. This is particularly worrisome because in fielded systems we will typically only have access to healthy samples. The transfer distance between the healthy source and target data suggests a much smaller change than actually occurs. This finding reaffirms our position that designing systems to avoid difficult transfer learning problems is essential to AI engineering because there are distributional changes over a system's life cycle that we cannot sample and empirically characterize in the field.

In PHM systems, it may be the case that some failure modes are similar across many machines or many rebuilds, whereas others are not. Transfer distance provides a means for empirically quantifying how transferable failure modes are relative to each other, and thereby serves as a mechanism for directing related engineering effort, such as data collection and algorithm design.

\begin{figure*}[t]
     \centering
     \begin{subfigure}[b]{0.32\textwidth}
         \centering
         \includegraphics[width=\textwidth]{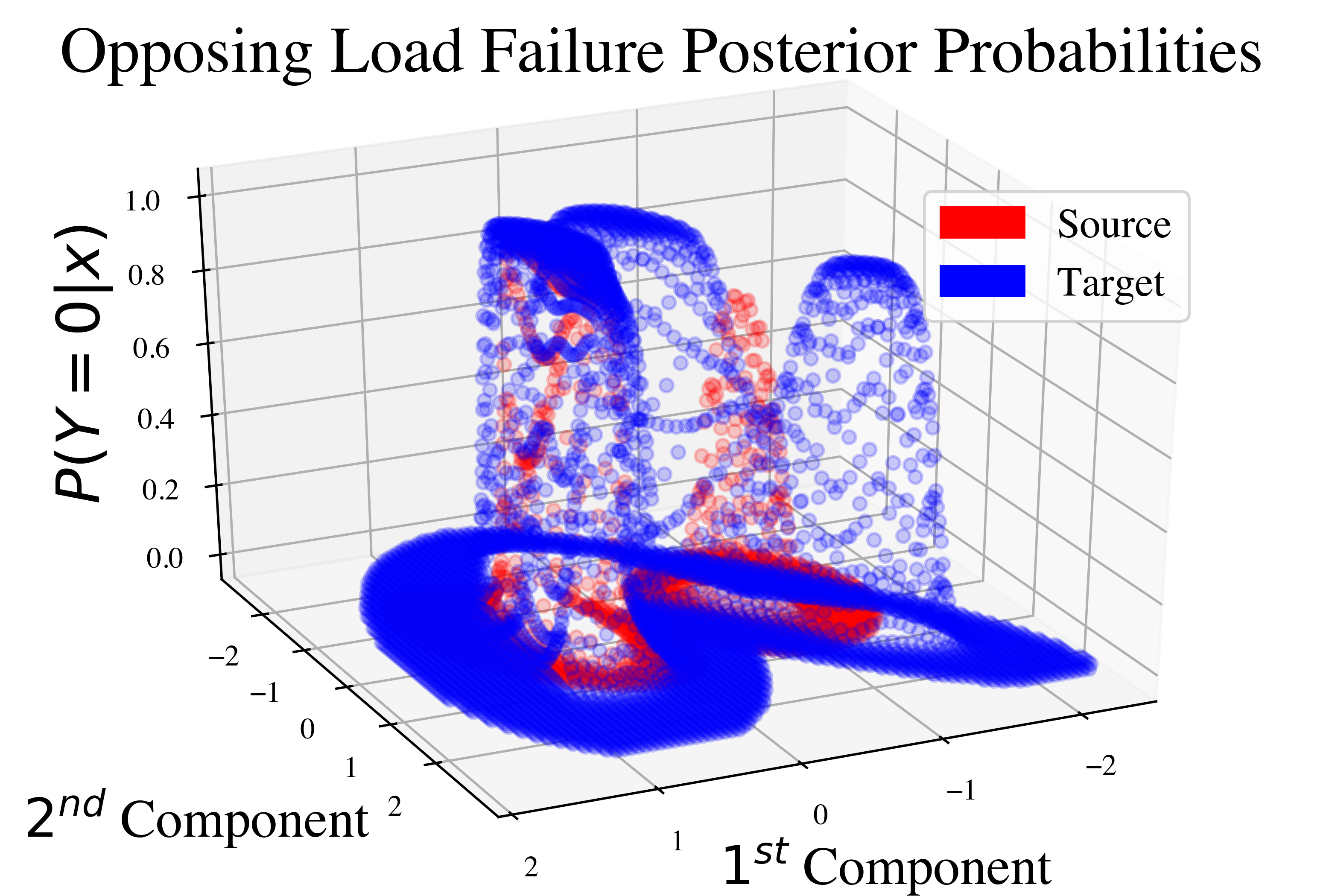}
         \caption{$p(X|Y=1)$}
         \label{fig:posterior-opposing}
     \end{subfigure}
     \hfill
     \begin{subfigure}[b]{0.32\textwidth}
         \centering
         \includegraphics[width=\textwidth]{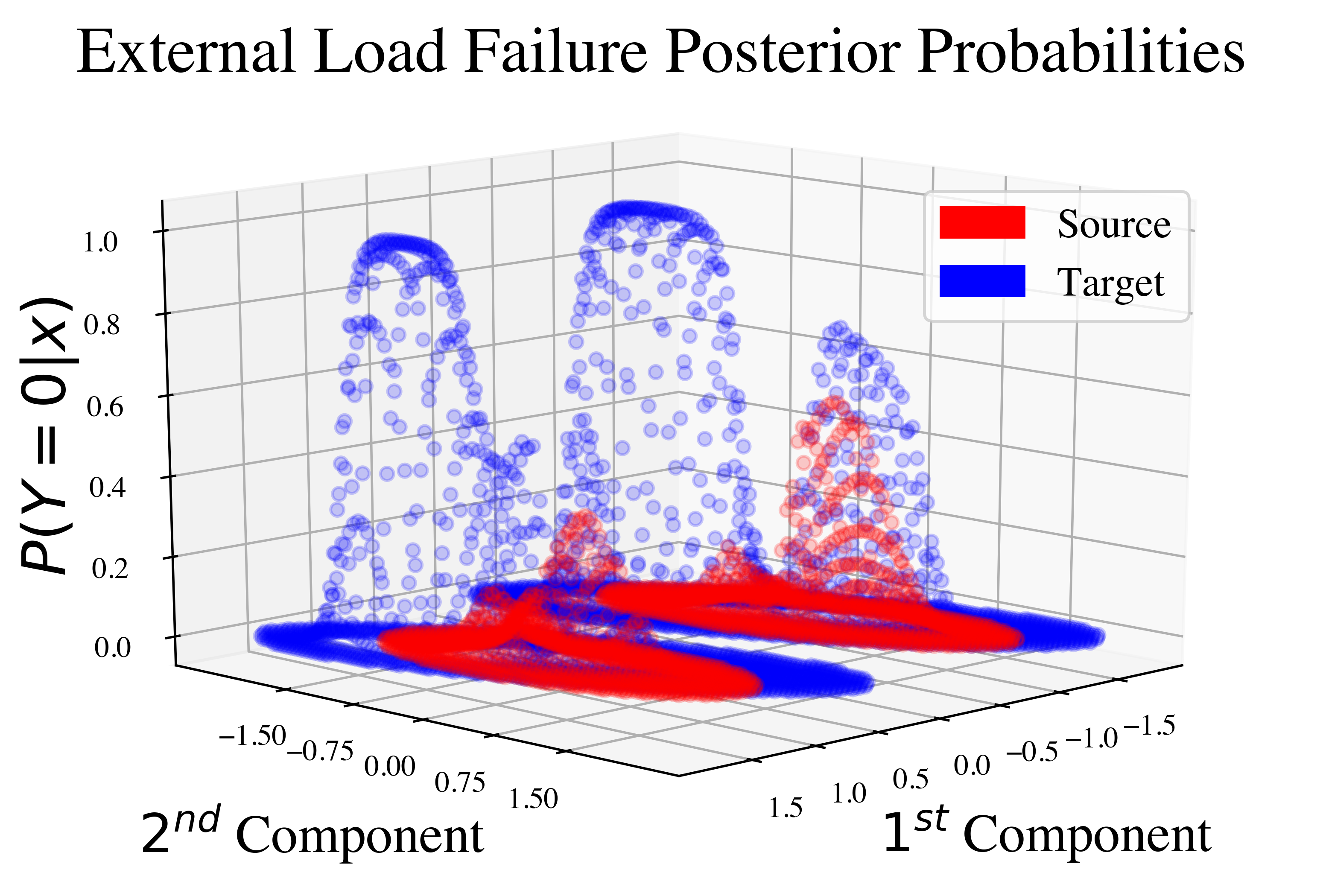}
         \caption{$p(X|Y=2)$}
         \label{fig:posterior-external}
     \end{subfigure}
     \hfill
     \begin{subfigure}[b]{0.32\textwidth}
         \centering
         \includegraphics[width=\textwidth]{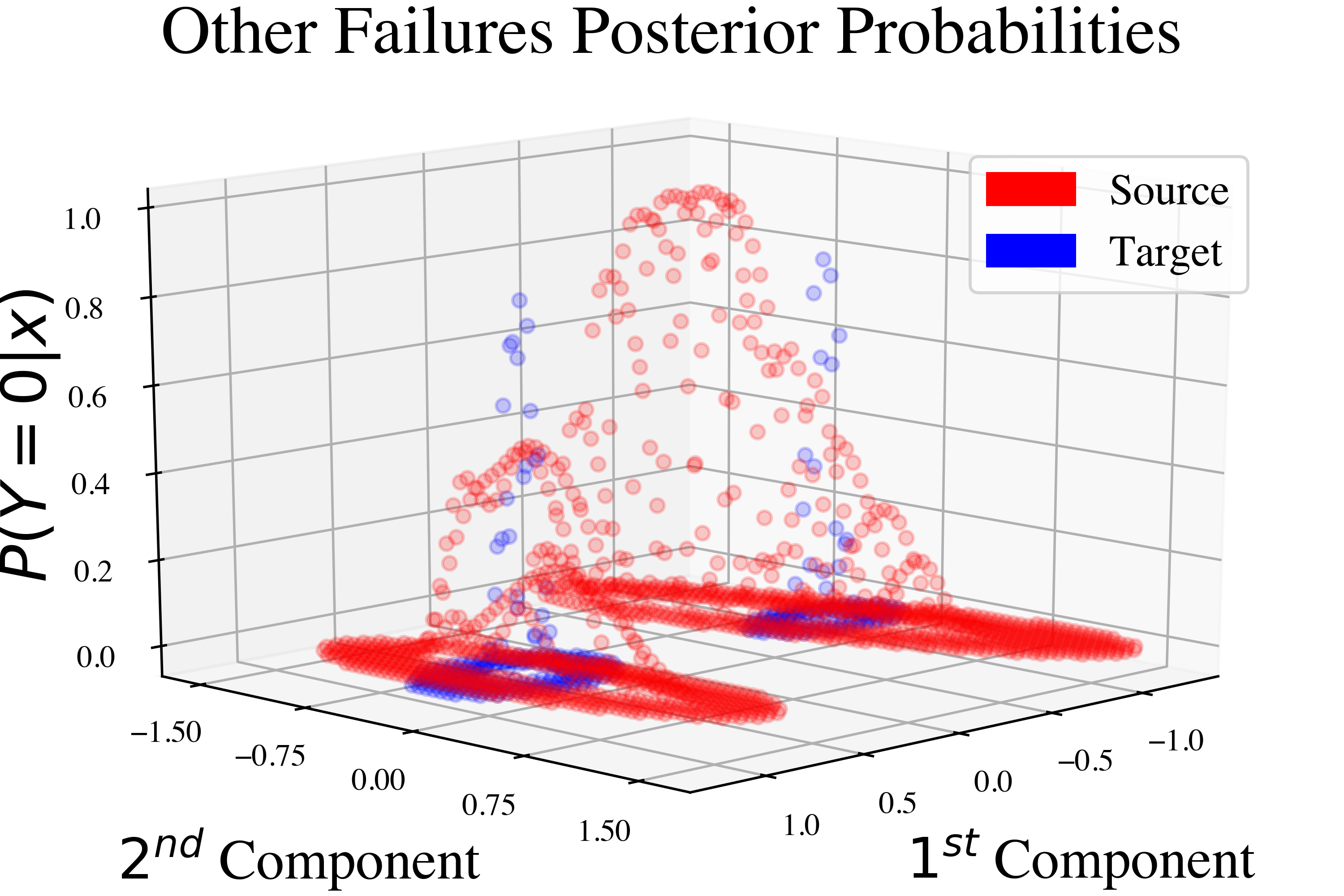}
         \caption{$P(Y=5|X)$}
         \label{fig:posterior-other}
     \end{subfigure}
        \caption{Posterior Distributions for Different Failure Modes}
        \label{fig:three-graphs-multiclass}
\end{figure*}

Since transfer learning comes with associated costs and risks, it is important to know where it is needed and where it is not. A need-based approach not only allows for reduced knowledge transfer and retraining, but also, it allows transfer learning algorithms to specifically focus on transferring knowledge for those failure modes which need source knowledge the most.

We quantify the transfer distance between failure modes in the source and target using a multi-class health state classification problem. Now, $\mathcal{Y} = \{0, 1, 2, 3, 4, 5\}$ where $Y=0$ signifies healthy and $Y = 1, ..., 5$ signify opposing load, external load, bypass valve, leak valve, and other failures, respectively. We have a similar number of samples between source and target and across failure modes. Using the presented methodology we fit a posterior distribution for $\forall y \in \mathcal{Y}$. Table \ref{table:multi} shows the likelihood and posterior transfer distances for each failure mode. 

Opposing load failures have a posterior transfer distance of 0.64 and leak valve failures have a posterior transfer distance of 0.88. This suggests that the sensor-data representations of opposing load failures in the source and target actuators are closer than those of leak valve failure. Put flatly, opposing load failures look more similar after the rebuild than leak valve failures. 

Figure \ref{fig:three-graphs-multiclass} shows the source and target posterior probabilities for opposing load, external load, and other miscellaneous failures. The overlap of the distributions in the plots corresponds to the posterior transfer distances in Table \ref{table:multi}. Perhaps an algorithm designer may conclude that knowledge transfer is feasible for opposing load failures, but not for other failures. Or, perhaps a systems engineer would suggest redesigning the rebuild procedure to bring those failure modes with a higher transfer distance closer in PCA space.

\begin{table}[t]
\centering
\ra{1.3}
\begin{tabular}{@{}lrr@{}}
\toprule
Failure Type & Likelihood $\delta_{X|Y}$ & Posterior $\delta_{Y|X}$ \\
\midrule
Opposing Load & 0.53 & 0.64 \\
External Load & 0.41 & 0.72 \\
Bypass Valve & 0.18 & 0.69 \\
Leak Valve & 0.74 & 0.88 \\
Other & 0.67 & 0.80 \\ 
\bottomrule
\\
\end{tabular}
\caption{Hellinger transfer distance for relevant distributions.}
\label{table:multi}
\end{table}

\subsection{Transfer Distance and Sample Size}

We have shown how transfer distance can be used to characterize transferability and provide insights for system and algorithm design. It is important to note that the distribution of the target actuator has a certain sample complexity. Transfer learning that relies on measures of distributional difference should wait for the distribution to settle first, otherwise methods such as sample weighting and selection will be using inaccurate estimates of distributional divergence. Similarly, transfer distance may require a number of samples to be collected before it can be considered a reliable metric for design and operational decision-making.

\begin{figure}[h]
    \centering
    \includegraphics[width=0.5\textwidth]{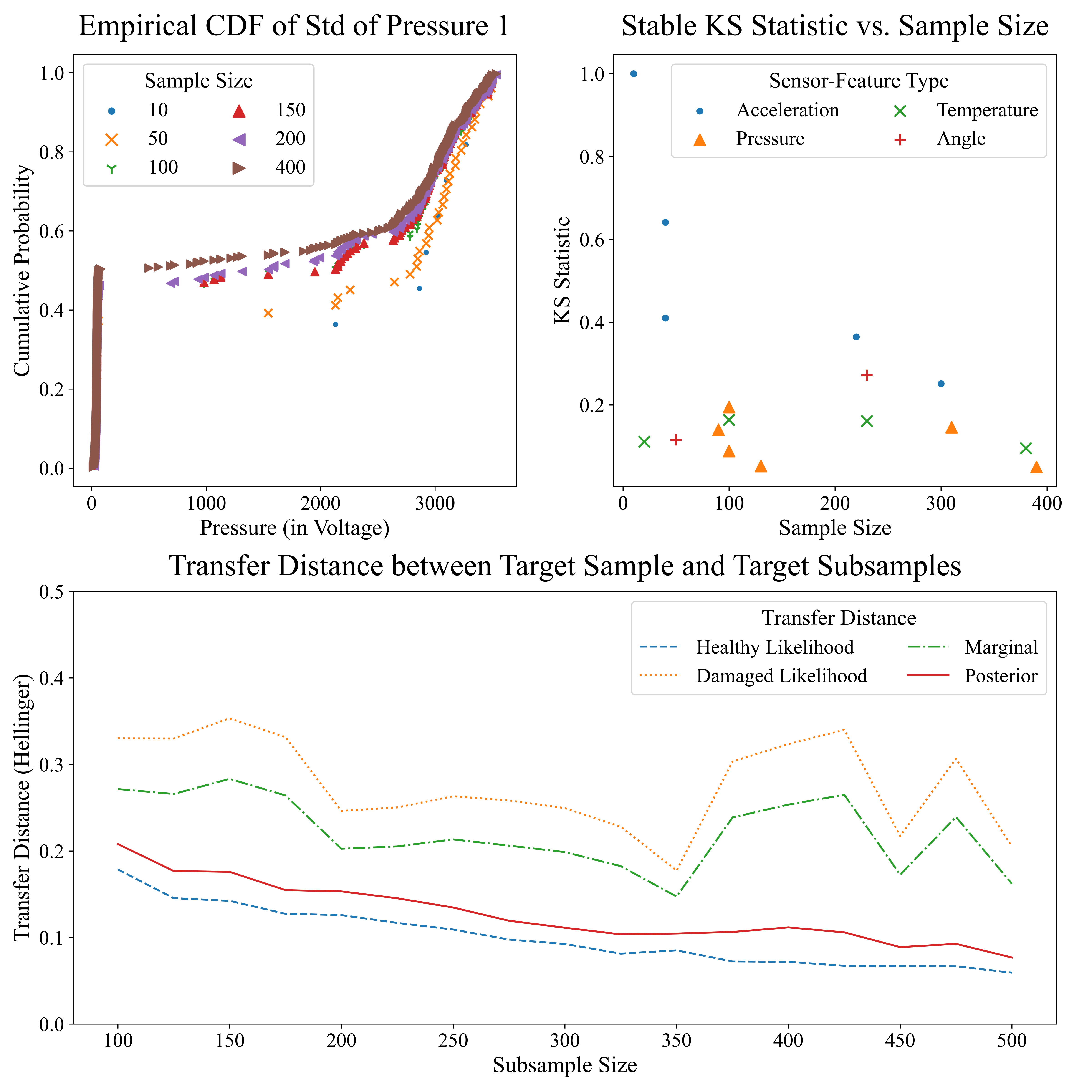}
    \caption{The top-left plot shows how the empirical cumulative distribution function (CDF) of the standard deviation of a pressure gauge changes with sample size. The top-right plot shows how many samples it takes for different sensor-feature types to converge to a stable value, labeled according to sensor-type. The bottom plot shows the transfer distance between GMMs trained on a subsample of target data and a full sample of target data.}
    \label{fig:sample-size}
\end{figure}

In the hydraulic actuators, each sensor-feature, e.g., the mean of acceleration 1, the standard deviation of pressure 1, etc., has its own sample complexity. Note the top-left plot in Figure \ref{fig:sample-size} which shows the empirical cumulative distribution functions (CDFs) associated with different size samples of the standard deviation of a pressure gauge. The CDF appears not to settle until 150 to 200 samples. If we use the Kolmogorov Smirnov (KS) statistic, which gives the largest absolute difference between two univariate CDFs, we can test when successive increases in sample size no longer change the distance between a sensor-feature's CDF in the source and target. In the top-right plot of Figure \ref{fig:sample-size} the point where the change in the KS statistic between the source and target for successive sample sizes changes less than 5\% is plotted for each type of sensor-feature, e.g., acceleration, pressure, etc. Apparently the distances between the source and target univariate CDFs converge at different rates. Accelerations have the largest KS statistics, but also the lowest sample size to settle.

We are learning using multiple sensor-features, thus, we are interested in how they settle jointly. In the bottom plot of Figure \ref{fig:sample-size} we consider sensor-feature interdependence by calculating the Hellinger transfer distances $\delta$ between target subsamples of a size corresponding to the x-axis and the full target sample. Transfer distances $\delta_{Y|X}$ and $\delta_{X|Y=0}$ decrease as sample size increases, and transfer distances $\delta_{X}$ and $\delta_{X|Y=1}$ roughly follow the same trend. Based on these results, it appears as though it takes at least 300 to 350 samples in the target before estimates of distributional divergence are stable. Note, that in practice, we often will only be able to conduct this analysis using healthy data.

In the context of machinery, depending on the nature of a maintenance procedure, the time to estimate the new distribution of sensor-data may change. This period relates to the lag-time before we can transfer knowledge to the new system to support data-driven PHM. The design of maintenance procedures to influence the length of this intervention is an important aspect of keeping PHM systems functioning.

\section{Transfer Distance for System Operation}

In system operation, transfer distance can be used to operate systems with an awareness of the expected generalization performance of component learning systems. Generalization performance concerns a learning system's error on new data. Different operational decisions are associated with different expected generalization performances. Inequality \ref{eq:1} suggests that transfer distance plays a fundamental role in determining the upper bound on error in new environments. Therefore, transfer distance has a strong connection to expected generalization performance.

In defense applications of computer vision, look angle, pixel density, time of day, and biome, for example, can vary between missions. Even when the sample spaces of images $\mathcal{X}$ and image labels $\mathcal{Y}$ have the same structure, the probability distributions associated with those sample spaces can differ drastically. Sometimes, one can intuit the existence of significant differences, for example, between image classification problems in the tundra and jungle. Other times, it is not as clear, for example, between classification problems in Southern and Northern California. In either case, transfer distance can empirically support or reject such intuition.

In the following, we first explore the relationships between transfer distance and expected generalization performance on the canonical handwritten digit recognition data set MNIST \cite{lecun1998mnist}. Then, with this understanding, we explore an application in defense where a model trained to detect the presence of aircraft in Southern California is deployed on a mission in Northern California \cite{kamsing2019deep}. In both cases we use auto-encoders to compress the images into a low-dimensional, latent representation before applying Algorithm \ref{algo:td} to compute transfer distances of interest. We use Gaussian mixture models as before, but now use KL divergence instead of Hellinger distance as our measure of transfer distance $\delta$.

\subsection{MNIST and Expected Operational Performance}

\begin{figure}[t]
    \centering
    \includegraphics[width=0.5\textwidth]{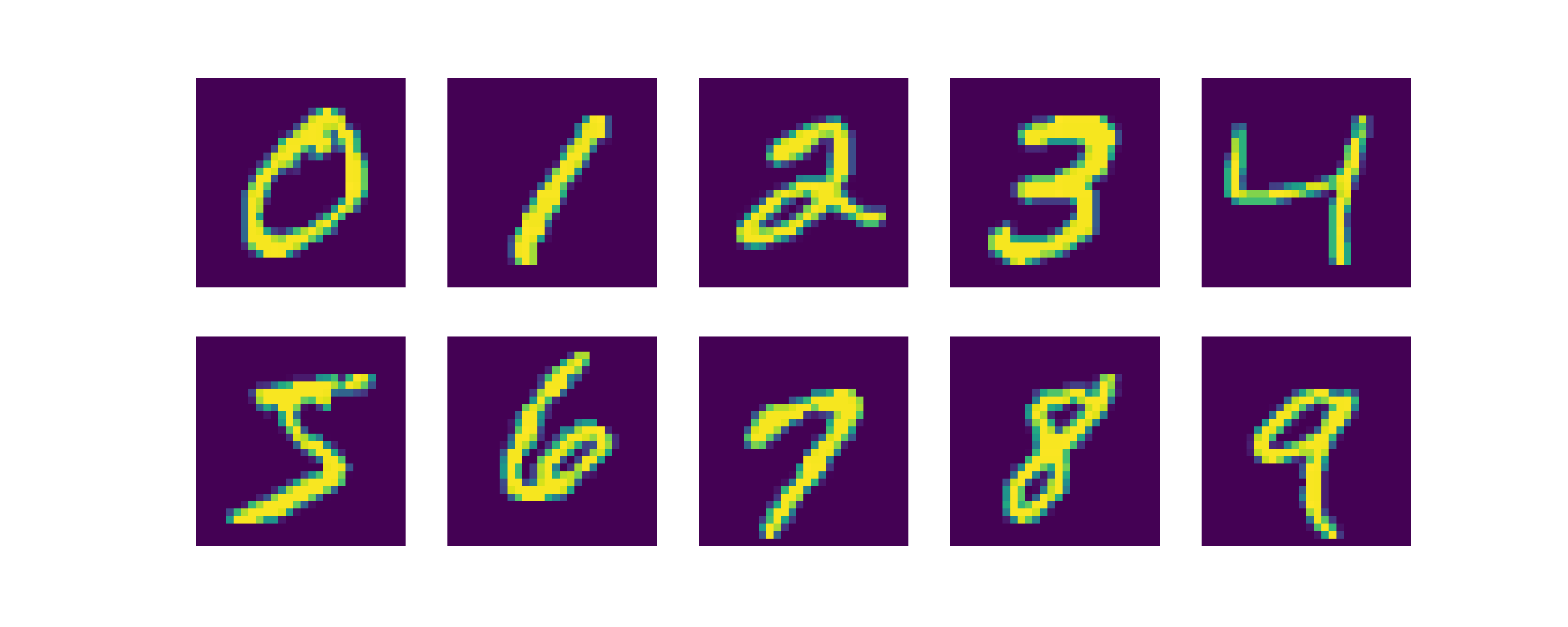}
    \caption{Example MNIST digit images 0-9 from left to right, top to bottom.}
    \label{fig:example-images}
\end{figure}

\begin{figure*}[t]
     \centering
     \begin{subfigure}[b]{0.42\textwidth}
         \centering
         \includegraphics[width=\textwidth]{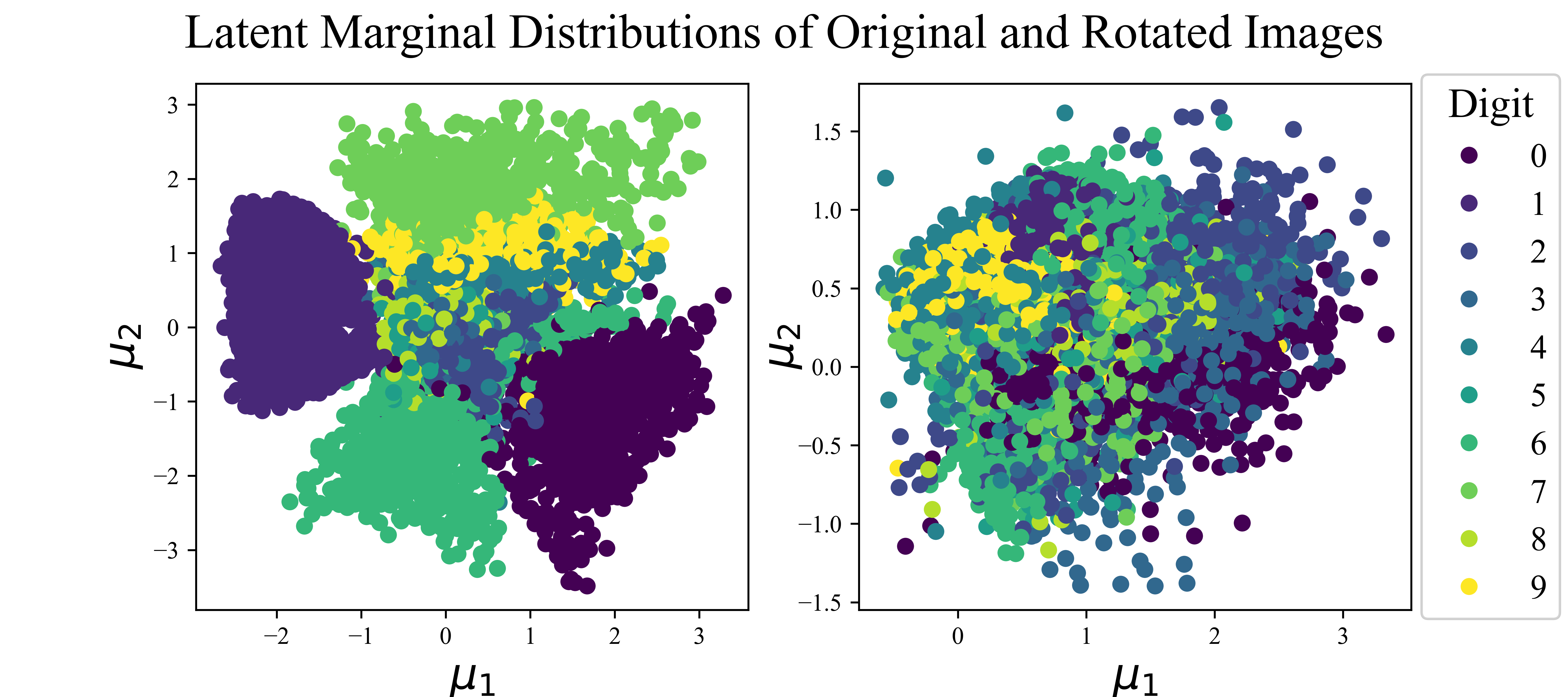}
         \caption{MNIST original, source images (left) and rotated, target images (right) in the variational auto-encoder's latent space.}
         \label{fig:MNIST-marginals}
     \end{subfigure}
     \hfill
     \begin{subfigure}[b]{0.57\textwidth}
         \centering
         \includegraphics[width=\textwidth]{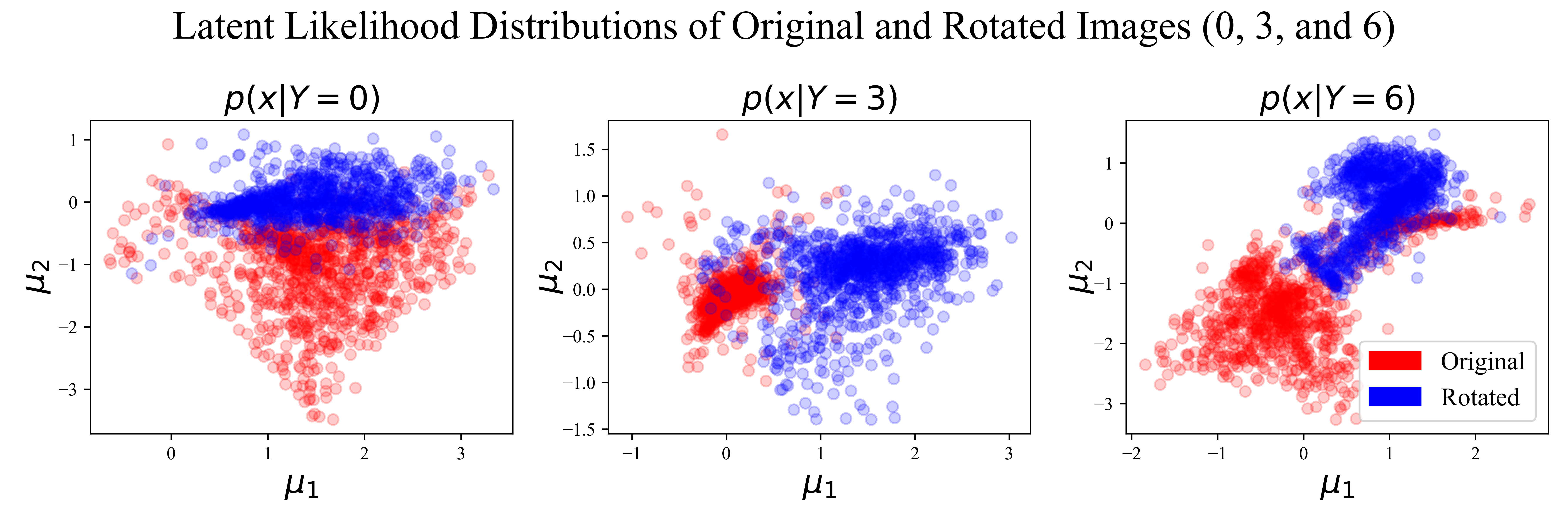}
         \caption{The variational auto-encoder's latent space shows the effect of rotation varies by digit.}
         \label{fig:MNIST-036}
     \end{subfigure}
     \caption{MNIST in Latent Space}
     \label{fig:mnist-td}
\end{figure*}

Just as transfer distance can be used as a metric for assessing the difficulty of generalization associated with a particular system design, it can be used to assess expected operational performance. Unlike in system design, in system operation we do not have direct control over transfer distance. We are not looking to change transfer distance directly, but rather, to operate in such a way that performance remains satisfactory \footnote{In general, design and operation are inextricable, but herein we establish a dichotomy to emphasize the dual use of transfer distance.}. Viewed discretely, we have a training environment, the source, and an operating environment, the target, and are interested in identifying if generalization performance in the operating environment will be satisfactory.

To see how transfer distance relates to expected operational performance, consider the MNIST handwritten digit recognition problem. The data set contains examples of handwritten digits 0 thru 9. We let the original data act as the source, training environment. To create a target we rotate all original data by 90 degrees clockwise. We fit a variational auto-encoder to the source images and use it to represent the source and target images as bivariate Gaussian distributions \cite{kingma2013auto}.

The transformed images are plotted in Figure \ref{fig:MNIST-marginals} according to their Gaussian means $\mu$. Whereas 0, 1, 6, and 7 are well separated in the source, as shown in the left plot, no rotated digits are well separated in the target, as shown in the right plot. The target images are interspersed with each other and have a smaller variance in $\mu_1$ and $\mu_2$ than the source images. This immediately suggests that the rotation of the images has a significant effect on $P(X)$.

This difference in $P(X)$ is not the same for all digits however. Consider the digits 0, 3, and 6, as shown in Figure \ref{fig:MNIST-036}. While all show differences, both the source and target `0' and `6' images share some overlap. In contrast, the source and target `3' images are almost partitioned by $\mu_1 = 0.5$. It makes intuitive sense that 0 and 6 are more similar because of the invariance of circles to rotation.

\begin{figure}[h]
    \centering
    \includegraphics[width=0.5\textwidth]{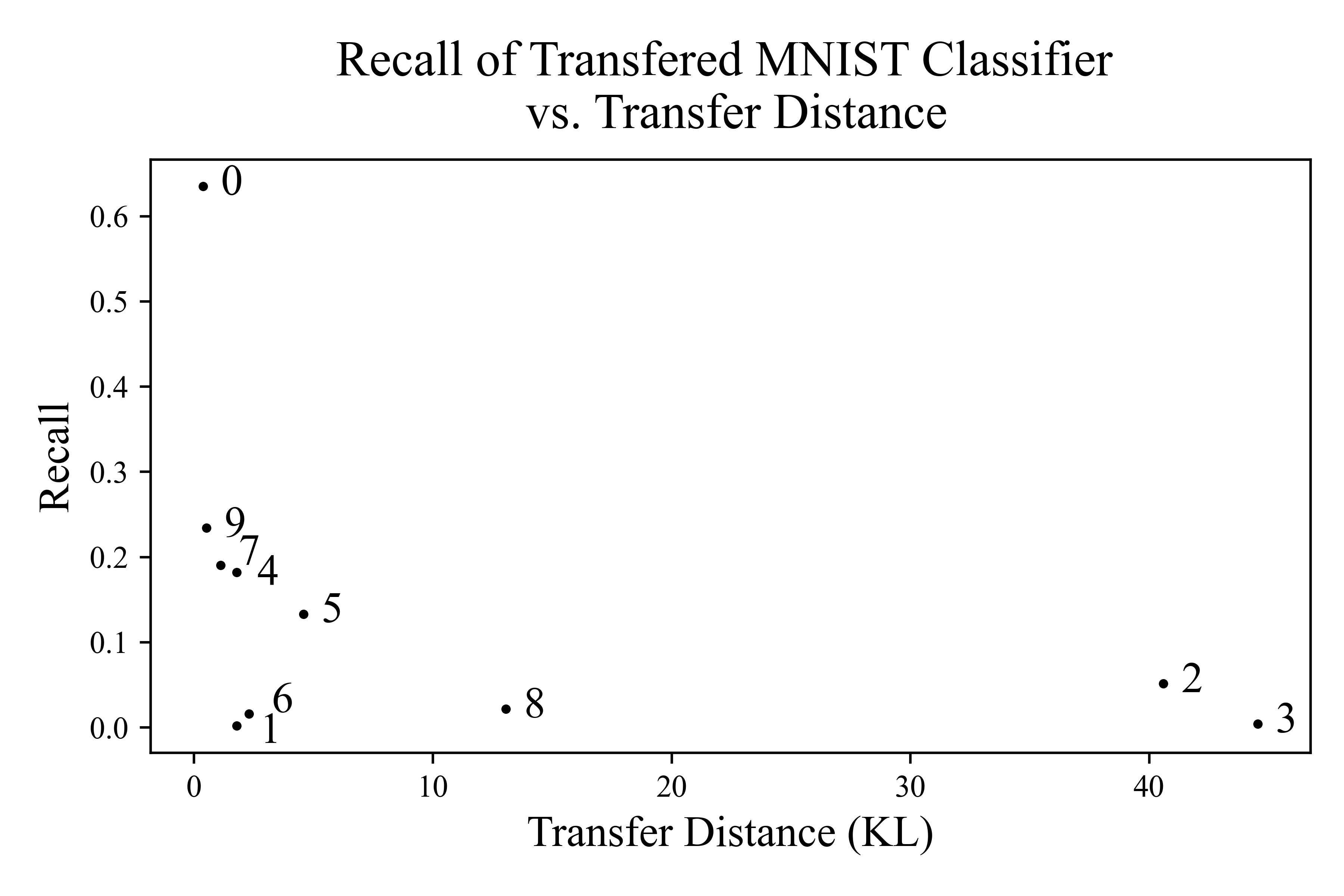}
    \caption{Higher recall digits tend to have lower transfer distance.}
    \label{fig:recall}
\end{figure}

To investigate further, we use a random forest to classify digits \cite{pal2005random}. When we calculate the recall on the rotated, target images of a classifier trained on the non-rotated, source images we find that those digits with a higher transfer distance (in this case a higher KL divergence) have a lower recall, as shown in Figure \ref{fig:recall}. Different to accuracy, recall considers the true positive rate, i.e., the ratio of correct classifications to number of instances of that class. `0' images have the highest recall and transfer distance, whereas `8' and `1' images have the lowest recall and transfer distance. Recall decreases with transfer distance. Given a measurement of transfer distance, we can form an empirical judgement of expected operational performance and, correspondingly, can make empirically informed operational decisions. In the following, we consider a `go, no go' mission deployment problem in aircraft detection. 

\subsection{Mission Scenario in Aircraft Detection}

\begin{figure}[t]
    \centering
    \includegraphics[width=0.5\textwidth]{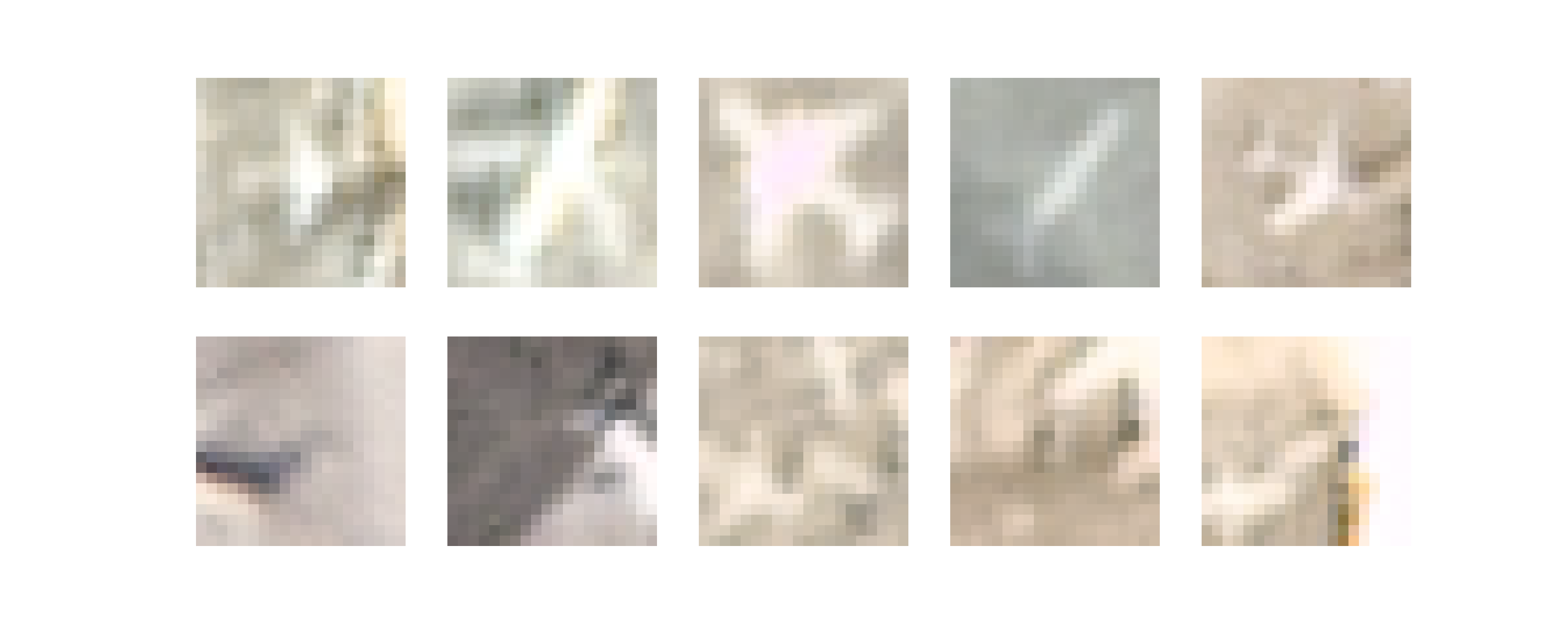}
    \caption{Example aircraft images and non-aircraft images from California in the top and bottom rows, respectively.}
    \label{fig:example-images2}
\end{figure}

Object detection from overhead imagery is a core function in defense systems. Despite the success of high-capacity models like deep learning, image classifiers are not global. Classifiers trained in one geographic region suffer performance degradation when deployed in other geographic regions. Fundamentally, this occurs because of a change in the underlying distribution of images. Transfer distance can be used to anticipate and detect drops in performance by comparing the distributional difference between samples from the training and operating environments.

Consider a case where a classifier is trained to detect the presence of aircraft in Southern California and is tasked with operating in Northern California. Example images are shown in Figure \ref{fig:example-images2}. There are roughly 20000 images from Southern California and 12000 images from Northern California. We trained a convolutional neural network to detect aircraft on Southern California images. 

\begin{figure}[h]
    \centering
    \includegraphics[width=0.5\textwidth]{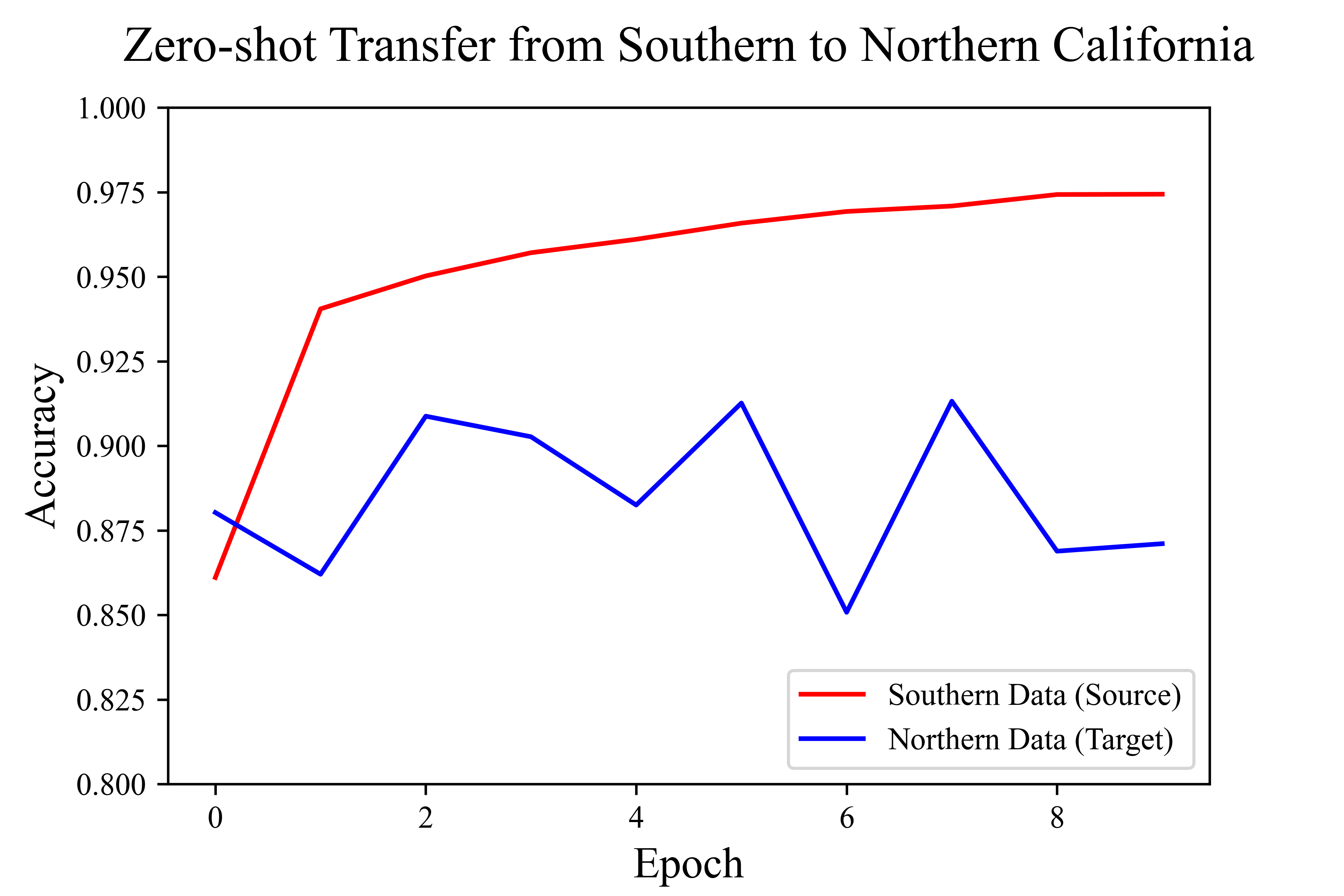}
    \caption{Shown is the classification accuracy of a convolutional neural network trained in Southern California evaluated on Southern California images, in blue, and on Northern California images, in orange, over the course of 10 training epochs.}
    \label{fig:north-to-south-error}
\end{figure}

When classifying held-out images from Southern California the classifier's accuracy is nearly 98\%, but when classifying images from Northern California accuracy drops to nearly 85\%, as shown in Figure \ref{fig:north-to-south-error}. The classifier still has predictive power, but, in critical applications like defense, the difference between a 2\% error rate and a 15\% error rate is significant enough to constitute failure. 

In order to apply our transfer distance methodology we first train an auto-encoder on the Southern California images. To do this, we initialize a convolutional auto-encoder with weights from the VGG-16 image classification network and then we fine-tune those weights \cite{simonyan2014very}. We use the auto-encoder to encode the images into vectors. Then, we find the principal components of the encoded Southern California images and transform all images into the first two principal components, as in the actuator example. In contrast to the actuator example, however, because of the size of the data set, we batch the data into samples of 100 before fitting Gaussian mixture models.

When we calculate transfer distances $\delta_X$ between samples drawn from Southern California, we find them to have a mean KL divergence of 5.60. When we calculate transfer distances $\delta_X$ between samples drawn from Southern and Northern California, we find them to have a slightly higher mean KL divergence of 5.97. This suggests that samples drawn from Southern and Northern California are, on average, farther from each other than two samples drawn from Southern California. The small difference in expected transfer distance corresponds to the slight drop in classification accuracy in Figure \ref{fig:north-to-south-error}.

We can investigate this trend by calculating transfer distances $\delta_{X|Y}$ of correctly and incorrectly classified images. Correctly classified Northern California aircraft images have a KL divergence of 0.94 from Southern California aircraft images, while misclassified Northern aircraft images have a KL divergence of 1.99, twice as high. These distances correspond to true positive and false negative cases, respectively. Correctly classified non-aircraft images from Northern California have a KL divergence of 2.34 from Southern California non-aircraft images, while misclassified non-aircraft images from Northern California have transfer distance of 3.11. Note, the transfer distance for incorrectly classified images is higher than the transfer distance for correctly classified images for both aircraft and non-aircraft images. That is, higher transfer distance correlates to higher error. We can analyze why this is so by using the principal components of the encoded images.

\begin{figure}[t]
     \centering
     \begin{subfigure}[b]{0.24\textwidth}
         \centering
         \includegraphics[width=\textwidth]{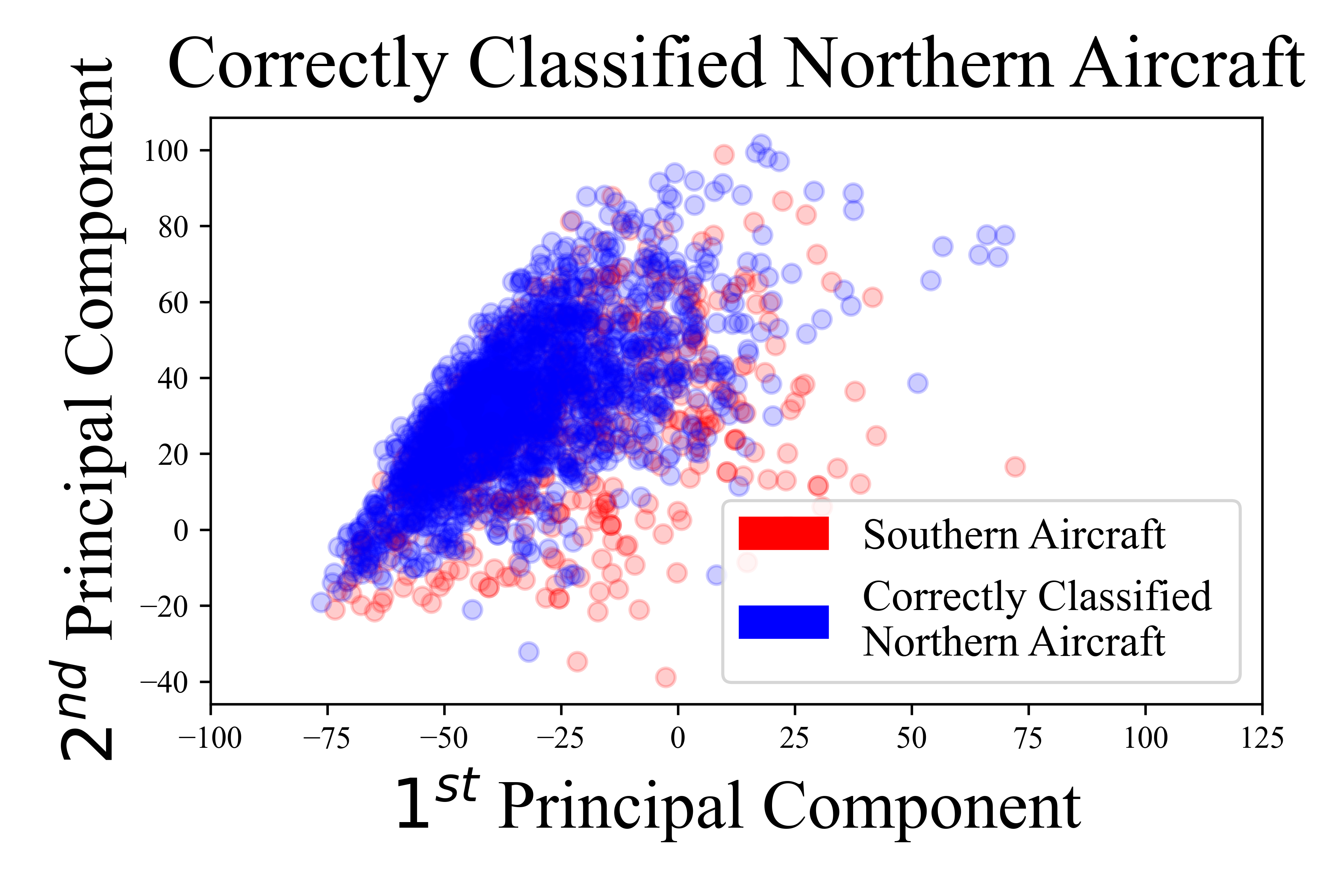}
         \caption{True positives}
         \label{fig:tp}
     \end{subfigure}
     \hfill
     \begin{subfigure}[b]{0.24\textwidth}
         \centering
         \includegraphics[width=\textwidth]{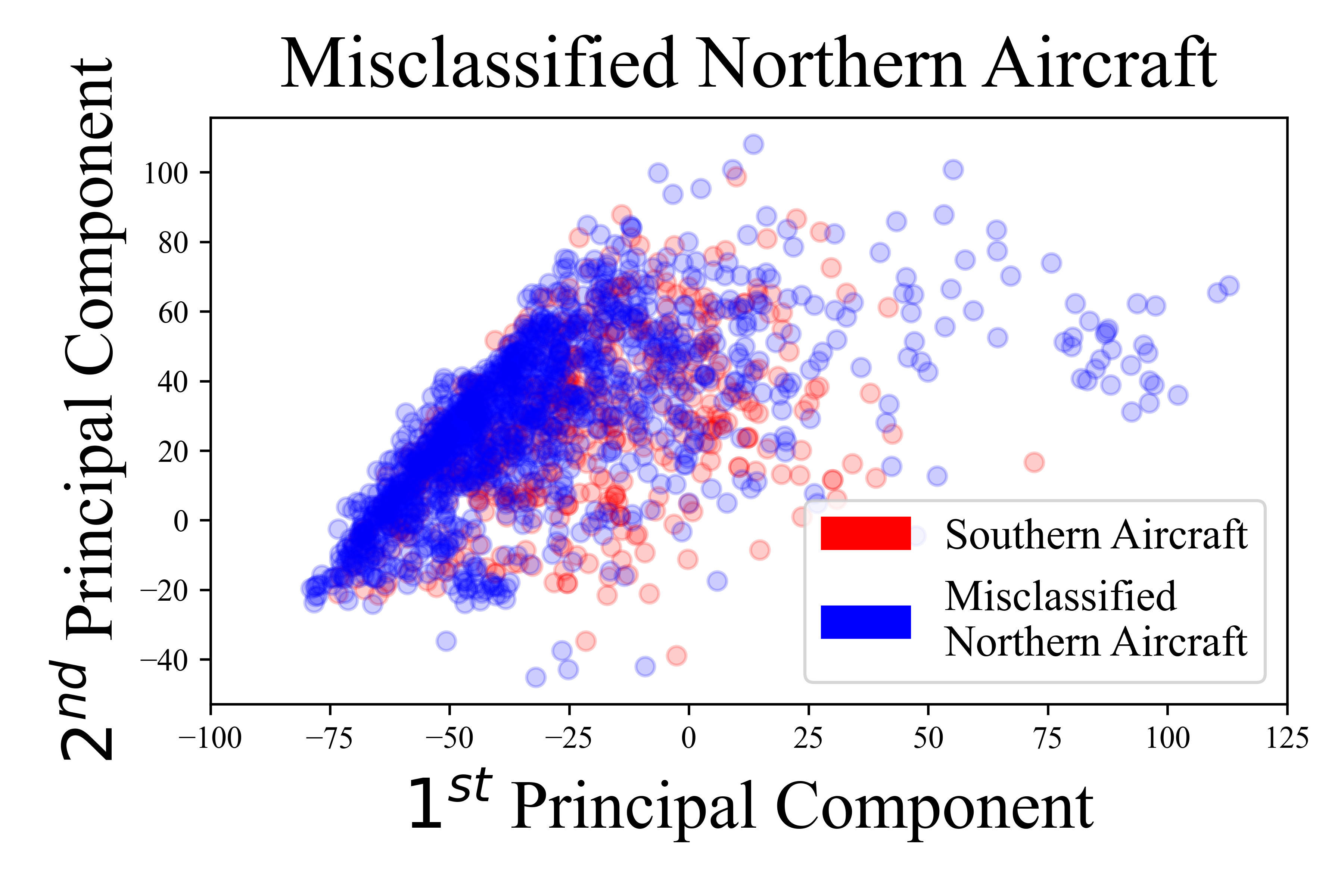}
         \caption{False negatives}
         \label{fig:fp}
     \end{subfigure}
     \hfill
    \caption{The first two principal components of true positives and false negative classifications when classifying images in Northern California using a classifier trained in Southern California. Misclassified Northern California aircraft images do not share a center of mass with Southern California aircraft images.}
    \label{fig:tp-and-fp}
\end{figure}

True positives refer to correctly classified aircraft images and false negatives refer to incorrectly classified aircraft images. The true positives and false negatives associated with the classifier trained on Southern California overhead imagery are shown in Figure \ref{fig:tp} and \ref{fig:fp}, respectively. Notice that the correctly classified aircraft images are near the center of mass of the Southern California aircraft images while the incorrectly classified aircraft images are not. In other words, the incorrectly classified Northern California aircraft images are in the tails of the distribution of Southern California aircraft images.

This suggests that system operators can empirically inform `go, no go' deployment decisions using transfer distance. In this case, the transfer distance between unlabeled images $\delta_X$ suggests a slight drop in performance. Further, transfer distance between misclassified images is higher than that of correctly classified images. Before deployment, system operators can use this empirical evidence to anticipate challenges to mission success. After deployment, system operators can use transfer distance to adjust their confidence in the model's classification accuracy in real-time.

\section{Conclusion}

As machine learning is deployed into systems, it is important to consider the role systems engineering plays as a mechanism for generalization. Systems engineering for AI requires metrics that can relate learning-theoretic concerns to the systems-level. Transfer distance is such a metric. In learning theory, it is central to the bounding of prediction error of learned models in new settings, such as rebuilt actuators or new look angles. At the systems-level, it serves as a measurement of the closeness of learning problems, and thereby a metric for designing and operating systems with the generalization performance of component learning systems in mind.

Herein, we formally defined transfer distance as a measure, presented an algorithm for calculating it, and demonstrated its use in system design and operation. We emphasized how, by using transfer distance as a metric, systems can be designed to influence generalization difficulty and can be operated to influence generalization performance.

We demonstrated how to use transfer distance to compare the transferability of binary and multi-class health state classification. In doing so, we showed how transfer distance can be used to quantify the transferability of both generalized and specific modes of failure across maintenance procedures. Also, we showed how to determine the number of samples needed for stable estimates of transfer distance and transfer learning parameters, and suggested the role of the design of maintenance procedures in the length of this intervening period. We also demonstrated transfer distance's use in computer vision. In particular, we identified which kinds of images are least transferable across changes in look angle and we anticipated and analyzed degradation in aircraft detection performance between geographic regions. We used different measures of transfer distance and generalization performance as well as different size data sets from different domains, i.e., sensor data and images, to highlight the generality of the methodology.

In future work, we plan to further explore the use of transfer distance in engineering practice. For example, in designing rebuild procedures, we aim to characterize the sensitivity of transfer distance to the tensions of fasteners, locations of sensors, and the manufacturer of replacement parts. Also, in making `go, no-go' operational decisions, e.g., in unmanned aerial systems, we aim to tie mission success to the transfer distance between training and operating environments.

\section{Acknowledgements}

The authors would like to thank Luna Innovations, Inc. for providing a portion of the data used within this paper. This material is based upon work supported by the Naval Sea Systems Command under Contract No. N00024-17-C4008. Any opinions, findings and conclusions or recommendations expressed in this material are those of the authors and do not necessarily reflect the views of the Naval Sea Systems Command.

\bibliographystyle{IEEEtran}
\bibliography{ref}

\begin{thebibliography}{10}
\providecommand{\url}[1]{#1}
\csname url@samestyle\endcsname
\providecommand{\newblock}{\relax}
\providecommand{\bibinfo}[2]{#2}
\providecommand{\BIBentrySTDinterwordspacing}{\spaceskip=0pt\relax}
\providecommand{\BIBentryALTinterwordstretchfactor}{4}
\providecommand{\BIBentryALTinterwordspacing}{\spaceskip=\fontdimen2\font plus
\BIBentryALTinterwordstretchfactor\fontdimen3\font minus
  \fontdimen4\font\relax}
\providecommand{\BIBforeignlanguage}[2]{{%
\expandafter\ifx\csname l@#1\endcsname\relax
\typeout{** WARNING: IEEEtran.bst: No hyphenation pattern has been}%
\typeout{** loaded for the language `#1'. Using the pattern for}%
\typeout{** the default language instead.}%
\else
\language=\csname l@#1\endcsname
\fi
#2}}
\providecommand{\BIBdecl}{\relax}
\BIBdecl

\bibitem{rogers2019adversarial}
C.~Rogers, J.~Bugg, C.~Nyheim, W.~Gebhardt, B.~Andris, E.~Heitman, and
  C.~Fleming, ``Adversarial artificial intelligence for overhead imagery
  classification models,'' in \emph{2019 Systems and Information Engineering
  Design Symposium (SIEDS)}.\hskip 1em plus 0.5em minus 0.4em\relax IEEE, 2019,
  pp. 1--6.

\bibitem{pan2009survey}
S.~J. Pan and Q.~Yang, ``A survey on transfer learning,'' \emph{IEEE
  Transactions on knowledge and data engineering}, vol.~22, no.~10, pp.
  1345--1359, 2009.

\bibitem{cody2019systems}
T.~Cody, S.~Adams, and P.~A. Beling, ``A systems theoretic perspective on
  transfer learning,'' in \emph{2019 IEEE International Systems Conference
  (SysCon)}.\hskip 1em plus 0.5em minus 0.4em\relax IEEE, 2019, pp. 1--7.

\bibitem{rosenstein2005transfer}
M.~T. Rosenstein, Z.~Marx, L.~P. Kaelbling, and T.~G. Dietterich, ``To transfer
  or not to transfer,'' in \emph{NIPS 2005 workshop on transfer learning}, vol.
  898, 2005, pp. 1--4.

\bibitem{wang2019characterizing}
Z.~Wang, Z.~Dai, B.~P{\'o}czos, and J.~Carbonell, ``Characterizing and avoiding
  negative transfer,'' in \emph{Proceedings of the IEEE Conference on Computer
  Vision and Pattern Recognition}, 2019, pp. 11\,293--11\,302.

\bibitem{jiang2008literature}
J.~Jiang, ``A literature survey on domain adaptation of statistical
  classifiers,'' \emph{URL: http://sifaka. cs. uiuc.
  edu/jiang4/domainadaptation/survey}, vol.~3, pp. 1--12, 2008.

\bibitem{blitzer2008learning}
J.~Blitzer, K.~Crammer, A.~Kulesza, F.~Pereira, and J.~Wortman, ``Learning
  bounds for domain adaptation,'' in \emph{Advances in neural information
  processing systems}, 2008, pp. 129--136.

\bibitem{ben2010theory}
S.~Ben-David, J.~Blitzer, K.~Crammer, A.~Kulesza, F.~Pereira, and J.~W.
  Vaughan, ``A theory of learning from different domains,'' \emph{Machine
  learning}, vol.~79, no. 1-2, pp. 151--175, 2010.

\bibitem{mohri2009rademacher}
M.~Mohri and A.~Rostamizadeh, ``Rademacher complexity bounds for non-iid
  processes,'' in \emph{Advances in Neural Information Processing Systems},
  2009, pp. 1097--1104.

\bibitem{zhang2012generalization}
C.~Zhang, L.~Zhang, and J.~Ye, ``Generalization bounds for domain adaptation,''
  in \emph{Advances in neural information processing systems}, 2012, pp.
  3320--3328.

\bibitem{webb2016characterizing}
G.~I. Webb, R.~Hyde, H.~Cao, H.~L. Nguyen, and F.~Petitjean, ``Characterizing
  concept drift,'' \emph{Data Mining and Knowledge Discovery}, vol.~30, no.~4,
  pp. 964--994, 2016.

\bibitem{wu2005tracking}
J.~Wu, X.-S. Hua, and B.~Zhang, ``Tracking concept drifting with gaussian
  mixture model,'' in \emph{Visual Communications and Image Processing 2005},
  vol. 5960.\hskip 1em plus 0.5em minus 0.4em\relax International Society for
  Optics and Photonics, 2005, p. 59604L.

\bibitem{diaz2018clustering}
J.~Diaz-Rozo, C.~Bielza, and P.~Larra{\~n}aga, ``Clustering of data streams
  with dynamic gaussian mixture models: an iot application in industrial
  processes,'' \emph{IEEE Internet of Things Journal}, vol.~5, no.~5, pp.
  3533--3547, 2018.

\bibitem{ditzler2011hellinger}
G.~Ditzler and R.~Polikar, ``Hellinger distance based drift detection for
  nonstationary environments,'' in \emph{2011 IEEE symposium on computational
  intelligence in dynamic and uncertain environments (CIDUE)}.\hskip 1em plus
  0.5em minus 0.4em\relax IEEE, 2011, pp. 41--48.

\bibitem{tsui2015prognostics}
K.~L. Tsui, N.~Chen, Q.~Zhou, Y.~Hai, and W.~Wang, ``Prognostics and health
  management: A review on data driven approaches,'' \emph{Mathematical Problems
  in Engineering}, vol. 2015, 2015.

\bibitem{li2008production}
J.~Li and S.~M. Meerkov, \emph{Production systems engineering}.\hskip 1em plus
  0.5em minus 0.4em\relax Springer Science \& Business Media, 2008.

\bibitem{delgado2012survey}
I.~R. Delgado, P.~J. Dempsey, and D.~L. Simon, ``A survey of current rotorcraft
  propulsion health monitoring technologies,'' 2012.

\bibitem{landolsi2017air}
F.~Landolsi, H.~Jammoussi, and I.~Makki, ``Air filter diagnostics \&
  prognostics in naturally aspired engines,'' in \emph{2017 IEEE International
  Conference on Prognostics and Health Management (ICPHM)}.\hskip 1em plus
  0.5em minus 0.4em\relax IEEE, 2017, pp. 61--65.

\bibitem{eftekhari2013online}
M.~Eftekhari, M.~Moallem, S.~Sadri, and M.-F. Hsieh, ``Online detection of
  induction motor's stator winding short-circuit faults,'' \emph{IEEE Systems
  Journal}, vol.~8, no.~4, pp. 1272--1282, 2013.

\bibitem{zhang2019data}
W.~Zhang, D.~Yang, and H.~Wang, ``Data-driven methods for predictive
  maintenance of industrial equipment: a survey,'' \emph{IEEE Systems Journal},
  vol.~13, no.~3, pp. 2213--2227, 2019.

\bibitem{shen2015bearing}
F.~Shen, C.~Chen, R.~Yan, and R.~X. Gao, ``Bearing fault diagnosis based on svd
  feature extraction and transfer learning classification,'' in
  \emph{Prognostics and System Health Management Conference (PHM), 2015}.\hskip
  1em plus 0.5em minus 0.4em\relax IEEE, 2015, pp. 1--6.

\bibitem{xie2016cross}
J.~Xie, L.~Zhang, L.~Duan, and J.~Wang, ``On cross-domain feature fusion in
  gearbox fault diagnosis under various operating conditions based on transfer
  component analysis,'' in \emph{Prognostics and Health Management (ICPHM),
  2016 IEEE International Conference on}.\hskip 1em plus 0.5em minus
  0.4em\relax IEEE, 2016, pp. 1--6.

\bibitem{lu2017deep}
W.~Lu, B.~Liang, Y.~Cheng, D.~Meng, J.~Yang, and T.~Zhang, ``Deep model based
  domain adaptation for fault diagnosis,'' \emph{IEEE Transactions on
  Industrial Electronics}, vol.~64, no.~3, pp. 2296--2305, 2017.

\bibitem{zhang2017new}
W.~Zhang, G.~Peng, C.~Li, Y.~Chen, and Z.~Zhang, ``A new deep learning model
  for fault diagnosis with good anti-noise and domain adaptation ability on raw
  vibration signals,'' \emph{Sensors}, vol.~17, no.~2, p. 425, 2017.

\bibitem{li2020fault}
X.~Li, Y.~Hu, M.~Li, and J.~Zheng, ``Fault diagnostics between different type
  of components: A transfer learning approach,'' \emph{Applied Soft Computing},
  vol.~86, p. 105950, 2020.

\bibitem{adams2017comparison}
S.~Adams, R.~Meekins, P.~A. Beling, K.~Farinholt, N.~Brown, S.~Polter, and
  Q.~Dong, ``A comparison of feature selection and feature extraction
  techniques for condition monitoring of a hydraulic actuator,'' in
  \emph{Annual Conference of the Prognostics and Health Management Society
  2017}, 2017.

\bibitem{meekins2018cost}
R.~Meekins, S.~Adams, P.~A. Beling, K.~Farinholt, N.~Hipwell, A.~Chaudhry,
  S.~Polter, and Q.~Dong, ``Cost-sensitive classifier selection when there is
  additional cost information,'' in \emph{International Workshop on
  Cost-Sensitive Learning}, 2018, pp. 17--30.

\bibitem{farinholt2018developing}
K.~M. Farinholt, A.~Chaudhry, M.~Kim, E.~Thompson, N.~Hipwell, R.~Meekins,
  S.~Adams, P.~Beling, and S.~Polter, ``Developing health management strategies
  using power constrained hardware,'' in \emph{PHM Society Conference},
  vol.~10, no.~1, 2018.

\bibitem{adams2019hierarchical}
S.~Adams, R.~Meekins, P.~A. Beling, K.~Farinholt, N.~Brown, S.~Polter, and
  Q.~Dong, ``Hierarchical fault classification for resource constrained
  systems,'' \emph{Mechanical Systems and Signal Processing}, vol. 134, p.
  106266, 2019.

\bibitem{adams2016condition}
S.~Adams, P.~A. Beling, K.~Farinholt, N.~Brown, S.~Polter, and Q.~Dong,
  ``Condition based monitoring for a hydraulic actuator,'' in \emph{Annual
  Conference of the Prognostics and Health Management Society October 2016},
  2016.

\bibitem{cody2019transferring}
T.~Cody, S.~Adams, P.~A. Beling, S.~Polter, K.~Farinholt, N.~Hipwell,
  A.~Chaudhry, K.~Castillo, and R.~Meekins, ``Transferring random samples in
  actuator systems for binary damage detection,'' in \emph{2019 IEEE
  International Conference on Prognostics and Health Management (ICPHM)}.\hskip
  1em plus 0.5em minus 0.4em\relax IEEE, 2019, pp. 1--7.

\bibitem{xiao2016overview}
X.~Xiao, D.~Xu, and W.~Wan, ``Overview: Video recognition from handcrafted
  method to deep learning method,'' in \emph{2016 International Conference on
  Audio, Language and Image Processing (ICALIP)}.\hskip 1em plus 0.5em minus
  0.4em\relax IEEE, 2016, pp. 646--651.

\bibitem{nanni2017handcrafted}
L.~Nanni, S.~Ghidoni, and S.~Brahnam, ``Handcrafted vs. non-handcrafted
  features for computer vision classification,'' \emph{Pattern Recognition},
  vol.~71, pp. 158--172, 2017.

\bibitem{goodfellow2016deep}
I.~Goodfellow, Y.~Bengio, A.~Courville, and Y.~Bengio, \emph{Deep
  learning}.\hskip 1em plus 0.5em minus 0.4em\relax MIT press Cambridge, 2016,
  vol.~1, no.~2.

\bibitem{chen2018end}
Z.~Chen, T.~Zhang, and C.~Ouyang, ``End-to-end airplane detection using
  transfer learning in remote sensing images,'' \emph{Remote Sensing}, vol.~10,
  no.~1, p. 139, 2018.

\bibitem{bulusu2020anomalous}
S.~Bulusu, B.~Kailkhura, B.~Li, P.~Varshney, and D.~Song, ``Anomalous instance
  detection in deep learning: A survey,'' Lawrence Livermore National
  Lab.(LLNL), Livermore, CA (United States), Tech. Rep., 2020.

\bibitem{shen2017wasserstein}
J.~Shen, Y.~Qu, W.~Zhang, and Y.~Yu, ``Wasserstein distance guided
  representation learning for domain adaptation,'' \emph{arXiv preprint
  arXiv:1707.01217}, 2017.

\bibitem{pan2008transfer}
S.~J. Pan, J.~T. Kwok, Q.~Yang \emph{et~al.}, ``Transfer learning via
  dimensionality reduction.'' in \emph{AAAI}, vol.~8, 2008, pp. 677--682.

\bibitem{long2017deep}
M.~Long, H.~Zhu, J.~Wang, and M.~I. Jordan, ``Deep transfer learning with joint
  adaptation networks,'' in \emph{International conference on machine
  learning}.\hskip 1em plus 0.5em minus 0.4em\relax PMLR, 2017, pp. 2208--2217.

\bibitem{tzeng2015simultaneous}
E.~Tzeng, J.~Hoffman, T.~Darrell, and K.~Saenko, ``Simultaneous deep transfer
  across domains and tasks,'' in \emph{Proceedings of the IEEE International
  Conference on Computer Vision}, 2015, pp. 4068--4076.

\bibitem{ganin2016domain}
Y.~Ganin, E.~Ustinova, H.~Ajakan, P.~Germain, H.~Larochelle, F.~Laviolette,
  M.~Marchand, and V.~Lempitsky, ``Domain-adversarial training of neural
  networks,'' \emph{The Journal of Machine Learning Research}, vol.~17, no.~1,
  pp. 2096--2030, 2016.

\bibitem{lecun1998mnist}
Y.~LeCun, ``The mnist database of handwritten digits,'' \emph{http://yann.
  lecun. com/exdb/mnist/}, 1998.

\bibitem{kamsing2019deep}
P.~Kamsing, P.~Torteeka, and S.~Yooyen, ``Deep convolutional neural networks
  for plane identification on satellite imagery by exploiting transfer learning
  with a different optimizer,'' in \emph{IGARSS 2019-2019 IEEE International
  Geoscience and Remote Sensing Symposium}.\hskip 1em plus 0.5em minus
  0.4em\relax IEEE, 2019, pp. 9788--9791.

\bibitem{kingma2013auto}
D.~P. Kingma and M.~Welling, ``Auto-encoding variational bayes,'' \emph{arXiv
  preprint arXiv:1312.6114}, 2013.

\bibitem{pal2005random}
M.~Pal, ``Random forest classifier for remote sensing classification,''
  \emph{International journal of remote sensing}, vol.~26, no.~1, pp. 217--222,
  2005.

\bibitem{simonyan2014very}
K.~Simonyan and A.~Zisserman, ``Very deep convolutional networks for
  large-scale image recognition,'' \emph{arXiv preprint arXiv:1409.1556}, 2014.

\end{thebibliography}

\end{document}